\documentclass[sigplan,10pt]{acmart}

\usepackage{blindtext}
\usepackage{makecell}
\usepackage{tabularx}
\usepackage{booktabs}
\usepackage{tikz}
\usepackage{algorithm}
\usepackage{algorithmic}
\usepackage{mathtools}
\usepackage{filecontents}
\usepackage{xspace}
\usepackage{multirow,multicol}
\usepackage{multicol}
\usepackage{ifthen}
\usepackage{optidef}
\usepackage{wrapfig}
\usepackage{graphicx}
\usepackage{subcaption}
\usepackage{float}
\usepackage{subcaption}
\usepackage{enumitem}
\usepackage{listings}
\usepackage{color}
\usepackage{xcolor}
\usepackage{wrapfig}

\usepackage[small, compact]{titlesec}
\newcommand{\sys}[0]{TurboSpec\xspace}
\newcommand{\lily}[1]{\textcolor{blue}{Lily: #1}}
\newcommand{\ion}[1]{\textcolor{red}{Ion: #1}}
\newcommand{\woosuk}[1]{\textcolor{brown}{Woosuk: #1}}
\newcommand{\simon}[1]{\textcolor{orange}{Simon: #1}}

\newcommand{\kt}[1]{\textcolor{brown}{Kuntai: #1}}
\newcommand{\alvin}[1]{\textcolor{red}{Alvin: #1}}

\renewcommand\footnotetextcopyrightpermission[1]{}
\usepackage{enumitem}

\setcopyright{rightsretained}

\begin{document}

\author{Xiaoxuan Liu$^{\text{1}}$\enskip Jongseok Park$^{\text{1}}$\enskip  Langxiang Hu$^{\text{2}}$\enskip Woosuk Kwon$^{\text{1}}$\enskip Zhuohan Li$^{\text{1}}$\enskip Chen Zhang$^{\text{3}}$\enskip \\ Kuntai Du$^{\text{4}}$\enskip Xiangxi Mo$^{\text{1}}$\enskip Kaichao You$^{\text{1,3}}$\enskip Alvin Cheung$^{\text{1}}$\enskip Zhijie Deng$^{\text{5}}$\enskip Ion Stoica$^{\text{1}}$\enskip Hao Zhang$^{\text{2}}$}
\affiliation{\vspace{1mm} $^{\text{1}}$UC Berkeley \enskip $^{\text{2}}$UCSD \enskip $^{\text{3}}$Tsinghua University\enskip $^{\text{4}}$University of Chicago  \enskip $^{\text{5}}$SJTU \country{}}

\newcommand\todo[1]{\ifthenelse{\equal{\showcomments}{yes}}{{\color{red} TODO: #1}}{\ignorespaces}}

\renewcommand{\figureautorefname}{Fig.}

\date{}

\title[\sys: Closed-loop Speculation for LLM Serving]{\sys: Closed-loop Speculation Control System for Optimizing LLM Serving Goodput}

\begin{abstract}
Large Language Model (LLM) serving systems batch concurrent user requests to achieve efficient serving. 
However, in real-world deployments, such \textit{inter-request parallelism} from batching is often limited by external factors such as low request rates or memory constraints. Recent works focus on \textit{intra-request parallelism} from speculative decoding as a solution to this problem. Unfortunately, benefits from intra-request parallelism are often fragile, as speculative decoding causes overhead, and speculated tokens may miss. We observe that speculative decoding may \textit{degrade} LLM serving performance if added naively without tuning to the incoming requests and the speculation method. 
To alleviate the need for expert tuning and make speculative decoding more robust, we present \textit{\sys}, a speculation control system that automatically profiles the execution environment and utilizes a feedback-based algorithm to dynamically adjust the amount of intra-request parallelism in LLM serving. \sys predicts ``goodput''---the amount of successfully generated tokens--- to evaluate and adjust intra-request parallelism amount to that with the highest goodput in runtime.
We implement \sys on a real-world LLM serving system vLLM and demonstrate its effectiveness across diverse workloads and hardware configurations, providing consistent performance improvements across all test scenarios.

\end{abstract}


\maketitle
\pagestyle{plain}

\section{Introduction}
\label{sec:intro}

One of the biggest challenges to efficient LLM serving is the inherent sequential dependencies of LLMs~\cite{leviathan2023fast}. The autoregressive nature of LLMs requires sequential token generation, as each output token depends on all previously generated tokens. 
This inherent constraint limits LLMs to generating tokens one at a time, preventing full utilization of modern AI accelerators' parallel processing capabilities. Consequently, this leads to suboptimal hardware utilization, resulting in reduced throughput and increased costs.
Fortunately, large-scale serving systems handle many requests concurrently, which provides an opportunity to increase parallelism. In particular, many previous research~\cite{yu2022orca, kwon2023efficient} focus on batching concurrent requests and leveraging \textit{inter-request parallelism} to achieve high hardware utilization and throughput. 

\begin{figure}[t]
    \includegraphics[width=0.99\linewidth]{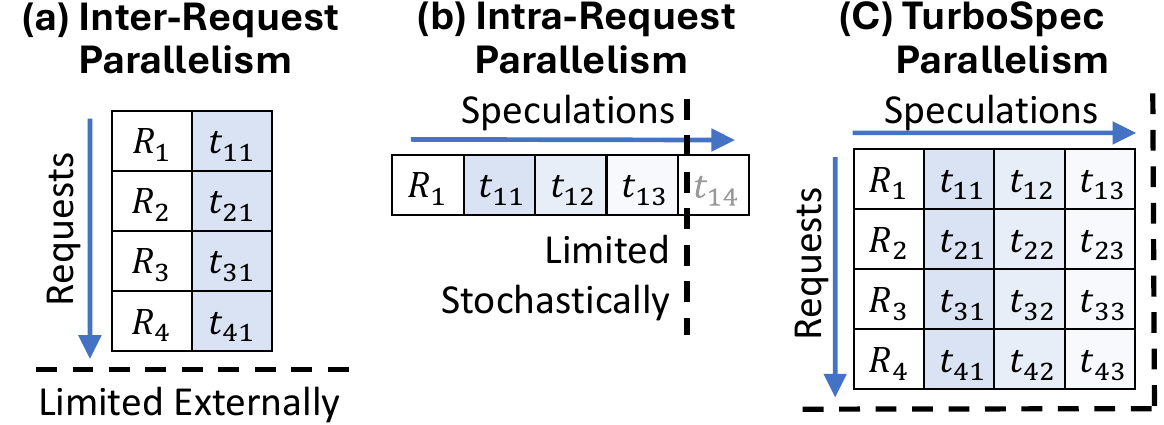}
    \vspace{-3pt}
    \caption{Comparison of (a) inter-request parallelism from batching, limited by external factors such as number of requests or VRAM, and (b) intra-request parallelism from speculative decoding, limited by stochastic factors such as the speculation method or the speculation difficulty of the request. \sys evaluates the limitations of (a) and (b) and adaptively combines them to efficiently leverage the larger product space of both parallelisms, as shown in (c).}
    \vspace{-10pt}
    \label{fig:parallelisms}
\end{figure}


However, in many real-world deployments, we find the \textit{inter-request parallelism} from request batching insufficient. For example, production environments often operate well below their maximum capacity to accommodate occasional bursts of requests~\cite{wang2024burstgpt}. Similarly, large LLMs and requests with large context lengths may leave little GPU memory available for batching any significant number of requests.  This raises the following questions: Are there alternatives to batching requests that allow us to exploit the parallelism of AI accelerators, such as GPUs, for LLM serving?



Fortunately, there is an orthogonal technique that leverages GPU parallelism even for a single request: \textbf{speculative decoding}~\cite{leviathan2023fast, chen2023accelerating}. Speculative decoding uses an efficient method, such as a small LLM model, to generate (speculate) multiple tokens ahead of time, and a larger model to verify these speculated tokens in \emph{parallel}. As such, speculative decoding enables \textit{intra-request parallelism}, where multiple tokens within a single request are processed (verified) in parallel, improving per-request latency. 
However, this method has an inherent limitation: tokens that fail verification must be discarded, resulting in wasted computational resources.
While the serving system can arbitrarily increase or decrease the number of speculated tokens (i.e., the intra-request parallelism), we cannot eliminate the wasted resources since perfect prediction accuracy cannot be guaranteed.


As a result, we cannot just simply integrate speculative decoding into LLM serving systems and hope for efficient resource usage. While computation inefficiencies due to incorrect token speculations may be acceptable under low system load, they become prohibitive if the acceptance rate drops or the request rate to the system increases. As explained later in Section~\ref{sec:motivation}, the efficacy of speculative decoding is influenced by many factors of the LLM serving system, such as batch size, hardware, models, or type of requests. Depending on the particular combination of these variables, speculative decoding could be beneficial or detrimental to system performance. This fragile nature of speculative decoding hints that it is infeasible for human operators to provide a “perfect” speculative decoding configuration that achieves the best performance for every LLM serving scenario.

This observation necessitates a new approach to integrating speculative decoding within LLM serving systems, guided by two key requirements:
\begin{itemize}
[nosep,leftmargin=1.5em,labelwidth=*,align=left]
 \item \textbf{Performance evaluation:} Provide a robust performance metric that can provide an abstraction for evaluating the speculative decoding performance across various speculation methods and serving environments.
 \item \textbf{Automatic Control:} Provide an automated system that optimizes the performance metric by maximally exploiting speculative decoding when it provides improvements and disabling speculative decoding when it does not.
 
\end{itemize}

\noindent To this end, we present \textbf{\sys}, a closed-loop control system for speculative decoding that adaptively adjusts the use of intra-request parallelism in LLM serving to optimize the serving system's \emph{goodput}. The goodput represents the number of correct tokens the system generates per unit time, i.e., it does not include the incorrect tokens generated by speculative decoding. Note that in the absence of speculative decoding, the goodput and throughput are the same.

\autoref{fig:parallelisms} illustrates our key insight into \sys. Comparing the inter-request parallelism of (a) and the intra-request parallelism of (b), we find that these two sources of parallelism are orthogonal and have complementary characteristics and limitations. By combining offline profiling and online feedback of the system operation, \sys is able to dynamically predict the impact of different intra-request parallelism configurations under the given execution environment as goodput. This allows \sys to adaptively mix speculative decoding and batching to maximize goodput: given a request rate, \sys batches requests to maximize parallelism and, if there are still available resources, uses speculative decoding to further improve goodput. 

We implement \sys on a production-grade serving system, vLLM~\cite{kwon2023efficient}, and perform extensive evaluations. 
Achieving accurate prediction of system goodput is difficult, as it would require a priori knowledge of the speculation outcome.
\sys tackles this challenge by implementing a two-part estimator of the goodput. First, it creates a performance model using an offline profiler to capture the LLM's execution latency across varying batch sizes and prediction lengths. Second, it uses an online predictor that leverages the performance model with historical token acceptance rate as feedback to forecast the expected number of tokens accepted for concurrent requests within each generation step. This leads to a robust goodput estimator,  
which serves as the foundation for optimizing our speculation length and verification strategy.

In summary, we make the following contributions:
\begin{itemize}
[nosep,leftmargin=1.5em,labelwidth=*,align=left]
    \item Identify the fragile nature of speculative decoding and highlight the necessity of an automated control system for speculative decoding in LLM serving.
    \item Design an automated speculative decoding control system for LLM serving, with definitions and insights into the key system metric goodput and a feedback-based speculation control algorithm based on goodput.
    \item Implement \sys, an adaptive speculation control system for LLM serving that dynamically optimizes speculation and provides robust and consistent performance improvements over different types of spec decode methods and serving scenarios.

\end{itemize}
\section{Background}
\label{sec:background}
\subsection{Batched LLM Serving Systems}
\label{sec:background-serving}

Given a list of tokens $(x_1, \ldots, x_n)$, a generative large language model (LLM)~\cite{bengio2000neural, achiam2023gpt} is trained to predict the conditional probability distribution for the next token: $P(x_{n+1} \mid x_1, \ldots, x_{n}).$ When deployed as a service~\cite{kwon2023efficient, pope2023efficiently}, LLMs take in a list of tokens from a user request and generate an output sequence $(x_{n+1}, \ldots, x_{n+T})$. The generation process requires evaluating the probability distribution and sampling the output token from the distribution, or decoding the output token, sequentially at every position for $T$ times. 

This sequential token-by-token generation limits the ability of generative LLMs to fully utilize the highly parallel AI accelerators such as GPUs~\cite{pope2023efficiently}.
Fortunately, in LLM serving scenarios, an abundant amount of concurrent inference requests are presumed available, leading many previous works~\cite{kwon2023efficient, zheng2024sglang} to leverage \textit{inter-request parallelism} from batching to solve the low parallelism problem. 
However, batching concurrent LLM requests is non-trivial: First, the requests may arrive at different times. A naive batching strategy would either make earlier requests wait for more requests or delay the incoming ones until the earlier ones finish, leading to queuing delays.
Second, the requests may have vastly different input and output token lengths, requiring padding the requests to equalize their lengths, wasting GPU computation and memory.

State-of-the-art LLM serving systems utilize continuous batching~\cite{gao2018low, yu2022orca} to address this problem. Instead of batching at the request level, continuous batching batches at each token generation step level. At each step, completed requests from the previous step are removed from the batch while the newly received requests are added, allowing the new request to immediately start its computation without queuing. This leads to a larger batch size and removes the need to pad the executions.
As such, this technique has been integrated within all popular LLM inference engines, such as vLLM~\cite{kwon2023efficient} and TensorRT-LLM~\cite{trtllm}.

\subsection{Speculative Decoding}
\label{sec:background-sd}


Although LLMs can only \textit{generate} output tokens sequentially, when provided with a list of pre-generated output tokens $x_{n+1}, \ldots, x_{n+T}$, LLMs can also efficiently \textit{evaluate} the probability distribution of each token $P(x_{n+1} \mid x_1, \ldots, x_{n}),$ $\ldots,$ $P(x_{n+T} \mid x_1, \ldots, x_{n + T - 1})$ in parallel. \emph{Speculative decoding}~\cite{leviathan2023fast, chen2023accelerating} leverages this property to enable \textit{intra-request parallelism}, using LLMs as an evaluator to decode a string of candidate tokens on a single request in parallel, potentially generating multiple tokens and reducing latency. 

For example, a smaller and more efficient draft model may propose a candidate token list $(x_{n+1}, \ldots, x_{n+k}),$ where $k$ is the number of candidates. Our target LLM model then predicts $k+1$ probability distributions $P(\cdot \mid x_1, \ldots, x_{n}), \ldots, P(\cdot  \mid x_1, \ldots, x_{n + k})$ in parallel. 
Based on the sampling method used, we accept a subset of tokens $x_{n+1}, \ldots, x_{n+m}$, where $m$ is the number of accepted tokens. 
A greedy sampling method, for instance, would check whether each $x_{n + i}$ is the token that maximizes the probability distribution $P(\cdot \mid x_1, \ldots, x_{n + i - 1})$ and accept the first $m$ tokens $x_{n+1}, \ldots, x_{n+m}$ that satisfy the condition. Additionally, for the position $n+m+1$, the correct token can always be sampled from the distribution $P(\cdot \mid x_1, \ldots, x_{n + m})$, as $x_{n+1}, \ldots, x_{n+m}$ are correct.

Therefore, each speculative decoding step generates $m+1$ tokens, with a \textbf{minimum of one token} ($m=0$) and a \textbf{maximum of \textit{k}+1 tokens}. Also, as the stochastic behavior of the target model's sampling process is untouched, the output of speculative decoding is identical to the non-speculative decoding output and has \textbf{zero accuracy loss}. 
There are two major approaches to generating the candidate tokens: (1) Draft LLM-based speculative decoding, which uses a smaller LLM as a draft model to generate the candidate tokens \cite{miao2023specinfer, liu2023online, zhou2023distillspec, cascade-inference, chen2024sequoia}. (2) Draft-model free speculative decoding, which uses either a branch of the target model or other sources (e.g., from an external database) to generate the candidate tokens \cite{cai2024medusa, li2024eagle, lin2024bita, fu2024break}. One example is prompt lookup decoding \cite{saxena2023prompt, somasundaram2024pld+}, which uses the best matching n-token string from the previous outputs as the speculated input. This effectively turns the generated output tokens into an n-gram language model and is often used in cases where the input is heavily repeated, such as code refactoring or grammar edits. 


Speculative decoding has been widely adopted as an optimization in real-world batched LLM serving systems (e.g., vLLM and TensorRT-LLM~\cite{trtllm2024specdecode}) and commercial offerings (e.g., OpenAI~\cite{openai2024predictedoutput} and AWS SageMaker~\cite{sagemaker2024specdecode}). However, these solutions require the system operators to manually specify the $k$ proposal length statically for each LLM deployment. Recent studies have shown that the optimal $k$ varies based on the current batch size~\cite{su2023synergy} and dataset~\cite{wang2024minions}. Choosing an incorrect $k$ can reduce throughput and increase system latency. As a result, while speculative decoding is a viable option, it has not become the default choice due to its rigidity and associated overhead, necessitating an automated way to efficiently enable and control speculative decoding in batched LLM serving systems.

\begin{figure}
    \centering
    \includegraphics[width=0.95\linewidth]{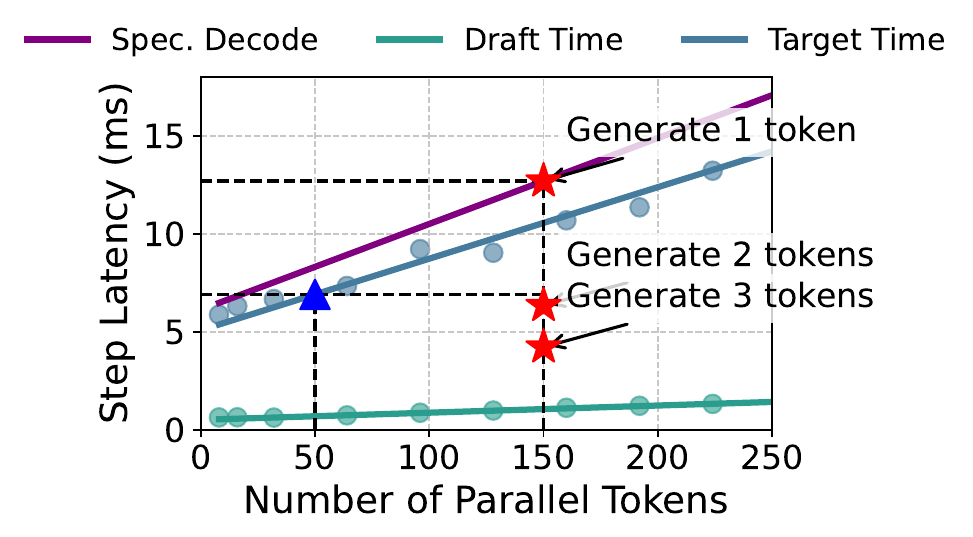}
    \vspace{-6pt}
    \caption{Single-step execution latency using vLLM~\cite{vllm2024serving} with Llama-2 7B model as the target model and Llama-2 160M model as the draft model on a H100 GPU. Blue triangle represents the latency without speculative decoding. Latencies are fitted from sample points, represented by the dots.}
    \label{fig:analysis_sd_with_batching_execution}
    \vspace{-10pt}
\end{figure}

\section{Challenges and Motivations}
\label{sec:motivation}
Before introducing system details, we first share insights into the characteristics of inter- and intra-request parallelism and its combination to understand what makes efficient speculative decoding in batched LLM serving systems challenging.
Based on these insights, we explain the motivations for our design choices in the construction of \sys. We start by comparing the core differences of the two parallelisms:

\noindent{\textbf{Availability:}} The main problem of inter-request parallelism is that it is dependent on the number of requests from the users, which is out of the serving system's control. In contrast, intra-request parallelism can be arbitrarily increased by the serving system using speculative decoding, which allows finer control of the parallelism in the system.

\noindent{\textbf{Usefulness:}} Intra-request parallelism from speculative decoding cannot be increased indefinitely, as more speculations generally mean more misses. This means that while the additional computation from batching is always useful, the additional computation from speculative decoding is only partially useful, where its usefulness is determined stochastically.

Therefore, while it is easy to imagine the convenience of having a system-controlled source of parallelism under low-traffic, tight SLO, or memory-bound scenarios with limited batching, it is hard to accurately perceive \textit{how useful} the added intra-request parallelism would be in such situations due to its stochastic properties.
To better understand the usefulness and drawbacks of combining intra- and inter-request parallelisms, we conduct a case study comparing a normal batched LLM execution against a toy speculative execution with a fixed amount of intra-request parallelism added per request. 

From \autoref{fig:analysis_sd_with_batching_execution}, we observe the non-speculative baseline depicted as a blue triangle achieving a latency of 7.4 ms with a batch size of 50. With speculation length $k = 2$, the number of tokens computed in parallel becomes $(k + 1) \times 50 = 150$, as explained in \autoref{sec:background-sd}. While the total speculative decoding latency (purple line) increases due to draft model overhead, the system now generates up to 3 tokens per request per step, with a minimum of one, yielding per-token latencies of 12.6 ms, 6.3 ms, and 4.2 ms for 1, 2, and 3 tokens respectively. Assuming an acceptance rate of 0.7, typical for Llama-based models, the probability of each case happening is 0.3, 0.21, 0.49, yielding a final expected latency of 7.16 ms.

We observe from the graph that intra-request parallelism of speculative decoding generates potential speedup by amortizing the static execution overhead, such as memory accesses, depicted by the y-intercept, across multiple tokens. However, as the rate of amortization is stochastically determined, it is uncertain how much, if any, speedup will be generated through the intra-request parallelism. If the serving system is too conservative in utilizing intra-request parallelism, it may lose potential speedup, and if it is too aggressive, it may degrade performance with a higher number of rejected tokens. 
Therefore, to achieve an efficient integration of speculative decoding, it is clear that LLM serving systems must be able to \textbf{accurately estimate} the performance impact of added intra-request parallelism on execution, a key factor that existing speculative decoding implementations miss.

However, achieving an accurate performance estimate is non-trivial as the speedups from intra-request parallelism are determined by various system metrics and configurations, as seen from \autoref{fig:analysis_sd_with_batching_execution}. To be specific, for the system to accurately calculate the expected speedup from intra-request parallelism, it must know (1) the latency profile of the original execution (blue line), (2) the latency profile of the speculative decoded execution (purple line), (3) the stochastic characteristics of the speculative decoding used, and (4) the batch size (x-axis) of the current step. Only after all these information are collected, the system can accurately calculate the expected speedup from intra-request parallelism and its probabilities to determine the most efficient use of speculative decoding.

However, we find that these execution characteristics are functions of three key factors of the system:
\begin{itemize}
[nosep,leftmargin=1.5em,labelwidth=*,align=left]
\item Acceptance Rate: Acceptance rate is the key metric to understand the stochastic property of speculative decoding. It is dynamically determined by the user request, the LLM model used, and the speculation method.
\item Request Rate: Request rate directly influences the number of batching available, and is dynamically determined by the number of users to the system and their usage patterns.
\item Hardware Usage: The available computation and memory resources determine the latency profile of the original execution and the speculative decoding execution. It also influences the maximum number of parallel tokens and the size of the LLM model.
\end{itemize}

Therefore, we conclude that for efficient utilization of speculative decoding in batched LLM serving systems, the system must be able to (1) understand the effects of the three key factors and model the behavior of speculative decoding under the given execution parameters to minimize the potential overhead and performance degradation from speculative decoding, 
and (2) accurately estimate the benefit of speculative decoding and dynamically adapt its use according to the estimation to extract the maximal benefit from intra-request and inter-request parallelism. 
We find utilizing a single comprehensive metric could greatly simplify the design of such a system. In \sys, we use \textit{goodput}, which estimates the system performance in the combined inter- and intra-request parallelism space, based on the three key factors. This effectively consolidates the stochastic properties of speculative decoding and the effects of the three key factors into a single quantitative measure, providing an efficient channel that simplifies the evaluation and the adaptation of speculative decoding in batched LLM serving scenarios, as further explained in the following section.
\section{Algorithm Design}
\label{sec:system}


\begin{figure*}[t]  
    \centering
    \includegraphics[width=0.999\linewidth]{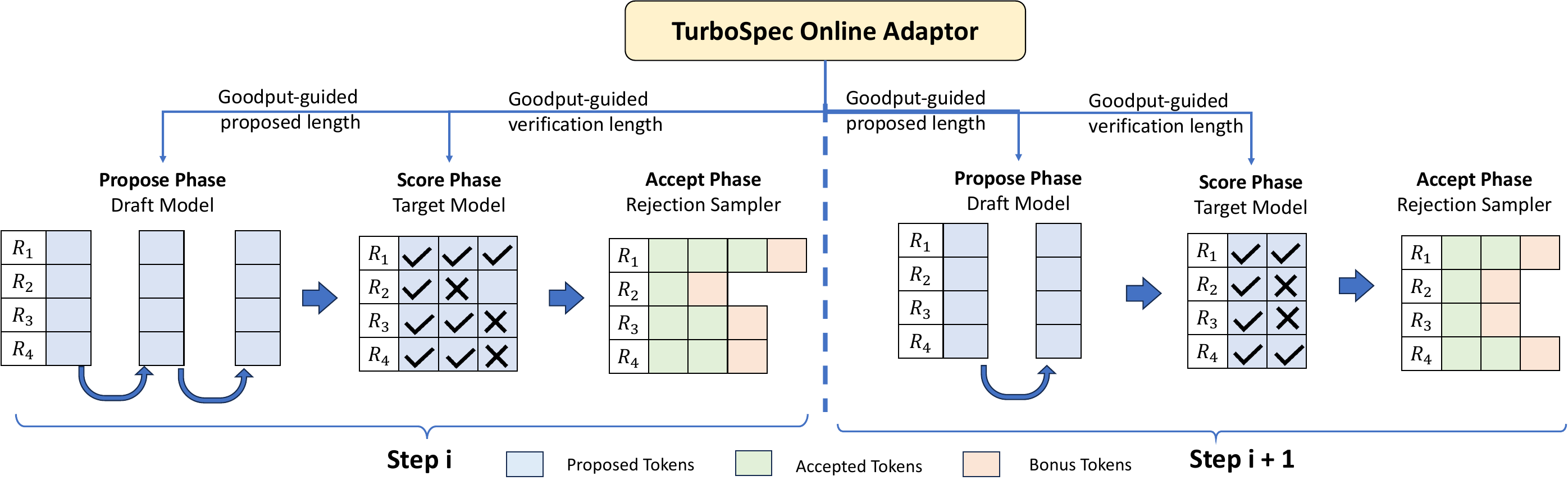}
    \caption{Two generation steps of draft-model based \sys execution. The proposed tokens from the draft model are sent to the target model for scoring in a single forward pass, allowing the generation of more than one token for each request. \sys online adaptor adjusts the proposed length and verification length dynamically for each step using goodput.}
    \label{fig:example}
    \vspace{-10pt}
\end{figure*}
\label{sec:background}

We first provide a high-level overview of our system design, showcasing how \sys integrates and controls speculative decoding in batched LLM executions using goodput as a key metric. We then provide a comprehensive description of goodput, followed by a detailed explanation of how \sys integrates with vLLM.

\autoref{fig:example} illustrates two consecutive generation steps of \sys which combines speculative decoding with continuous batching, managed by \sys Online Adaptor. Each generation step begins with proposing tokens autoregressively, followed by the target model scoring these proposals in a single forward pass. During these steps, our adaptor calculates goodput based on the key system metrics and optimizes both the proposed length of the drafting method and verification length of the target model using goodput.
Then, in the acceptance phase, we utilize rejection sampling to determine which tokens are retained. The final token sequence comprises two components: (1) draft-proposed and target-validated tokens, and (2) a bonus token that either rectifies an incorrect draft prediction or extends the sequence when all proposed tokens are accepted. After each step, the adaptor updates the acceptance rate based on the sampling result, to keep a accurate measurement of goodput across the execution.

As such, goodput is the key metric that determines the operation of \sys, and therefore must be carefully defined and managed through its operation. We provide the detailed definition, insights, and estimation process of goodput in the following secions.




\subsection{Using Goodput for Speculation}

\subsubsection{Automated Speculation with Goodput}
\label{sec:goodput}

As discussed in \autoref{sec:motivation}, the effectiveness of speculative decoding varies significantly due to acceptance rates from workload characteristics (dataset complexity, request patterns), user request rates (QPS), and hardware capabilities (e.g., GPU memory bandwidth, compute). Using a fixed proposed length across all configurations fails to capture these dynamics. For example, higher QPS may benefit from shorter proposals to better utilize inter-request parallelism, while lower QPS can favor longer proposals to exploit intra-request speculation. An adaptive mechanism is needed to  balance these tradeoffs and optimize the speculation degree.
While prior systems like vLLM~\cite{kwon2023efficient} and Orca~\cite{yu2022orca} focus on maximizing throughput (total output tokens per second), throughput is inadequate for speculative decoding when not all proposed tokens are accepted.


\noindent{\textbf{Goodput in Speculative Decoding.}}
In speculative decoding, not all tokens output by the target model are in the final output, as some tokens can be rejected.
To account for this, we define \textit{goodput} as:
\begin{equation}
    Goodput = \frac{Number\,of\,\textbf{Generated}\,Tokens}{Execution\,Time}
    \label{eq:goodput}
\end{equation}
i.e., goodput counts only the tokens that contribute to the final output: the validated speculative and bonus tokens. 

\begin{figure}[h]
    \centering
    \includegraphics[width=0.98\linewidth]{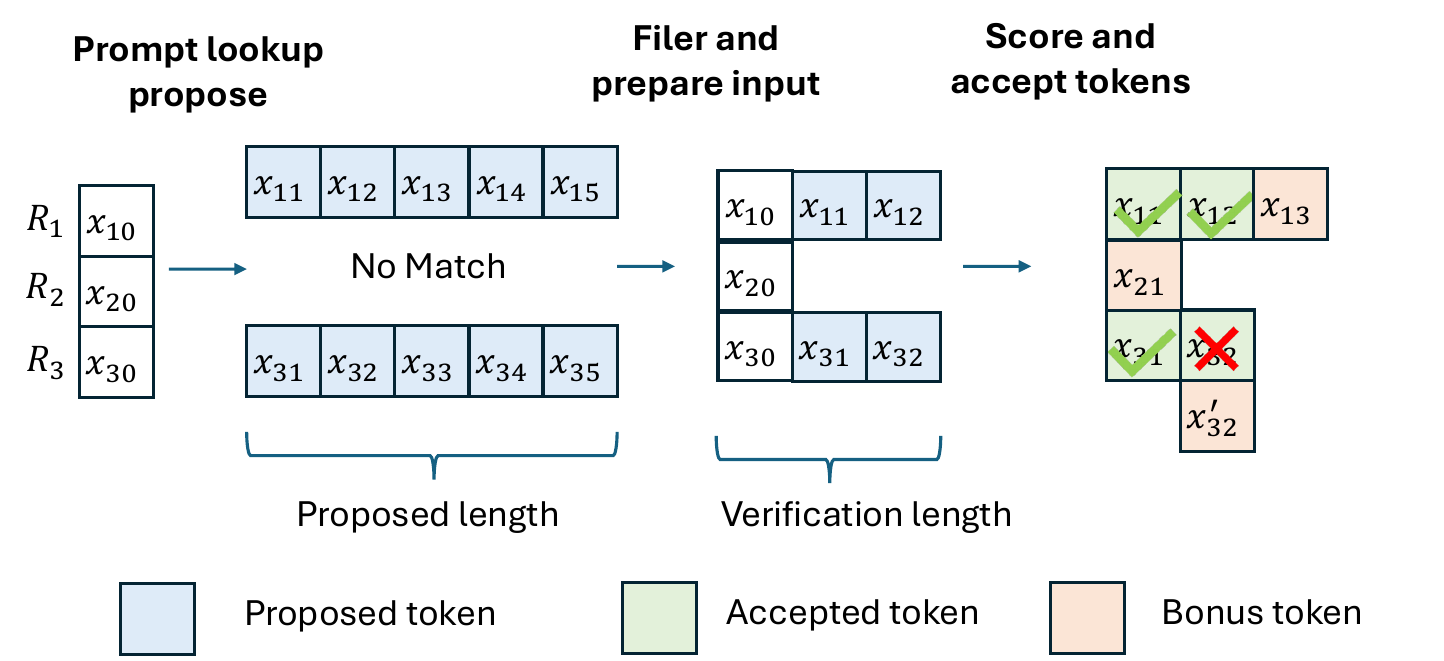}
    \caption{Example of different proposed and verification lengths in prompt lookup decoding across multiple requests. $x_{10}, x_{20}, x_{30}$ are input tokens for requests $R_1, R_2, R_3$ respectively. Successful matches are identified only for requests $R_1$ and $R_3$, resulting in proposals for these requests.}
    \label{fig:diff-propose-verify-len}
    \vspace{-10pt}
\end{figure}

\noindent{\textbf{Parameters of Goodput.}}
While the above definition is general across different speculative decoding algorithms, we focus on two configuration parameters: (1) Proposed length: the number of tokens proposed by the draft method in each step, and 
(2) Verification length: the number of tokens to be scored by the target method in each step.

\sys is designed so that the proposed length can differ from the generation length.
In prompt lookup decoding, the proposed and verification lengths may differ since not all requests can successfully retrieve tokens from the prompt. \autoref{fig:diff-propose-verify-len} gives one such example where $R_2$ does not have matched tokens during proposal. We use a fixed-length proposal strategy to handle this variability efficiently where each request attempts to retrieve a predetermined number of tokens. This approach is efficient for prompt lookup since the proposal cost only depends on the context search rather than the number of tokens proposed. The verification phase then processes only the successfully retrieved tokens.


\subsubsection{Understanding Goodput}
We next analyze how goodput guides speculative decoding decisions. Using profiling data from (Llama-160M draft, Llama2-7B target) on an H100 GPU, we examine goodput across different settings.
As seen from \autoref{fig:understand-goodput}, goodput has the following characteristics: 

\begin{figure}
    \centering
    \begin{subfigure}[t]{0.45\linewidth}
        \includegraphics[width=\linewidth]{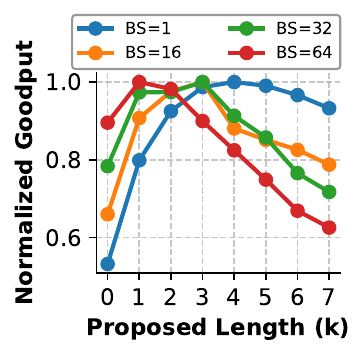}
        \caption{BS: Batch size.}
        \label{fig:goodput-bs}
    \end{subfigure}
    \hfill
    \begin{subfigure}[t]{0.45\linewidth}
        \includegraphics[width=\linewidth]{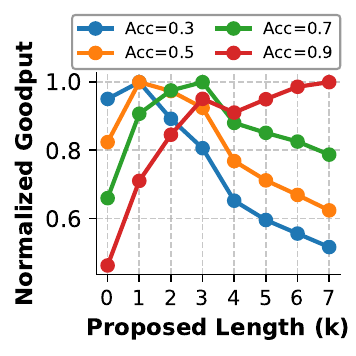}
        \caption{Acc: Acceptance rate.}
        \label{fig:goodput-acc}
    \end{subfigure}
    \vspace{-10pt}
    \caption{Normalized goodput across proposed lengths. We normalize goodput by expressing it as a fraction of the maximum goodput achieved across all proposed lengths in the same configuration (fix batch size or acceptance rate).}
    \label{fig:understand-goodput}
\end{figure}
\noindent{\textbf{(1) Propose more for small batches and less for large batches.}} \autoref{fig:goodput-bs} shows an inverse relationship between batch size and optimal proposed length. Small batches (BS=1) show peak goodput with longer proposals (k=4-5), while large batches (BS=64) perform best with shorter proposals (k=1-2). 

\noindent{\textbf{(2) Propose more for accurate batches and less for inaccurate batches.}}
\autoref{fig:goodput-acc} reveals that optimal proposed length varies with token acceptance rate. High acceptance rates (Acc=0.9) sustain strong goodput with longer proposals, while low rates (Acc=0.3) perform better with shorter proposals to minimize wasted computation.

\noindent\textbf{(3) Prefer batching over speculation.}
Consider a scenario where the acceptance of tokens is independent, with each token having a 0.7 probability of acceptance. In this case, the probability of accepting the first token is 0.7, while the probability of accepting both the first and second tokens is 0.7 $\times$ 0.7 = 0.49. Consequently, increasing the batch size tends to produce more tokens at the same cost. Doubling the batch size results in twice the number of generated tokens, whereas doubling the proposed length does not necessarily yield a proportional increase in output.


\subsubsection{Estimating Goodput}
\label{sec:estimate-goodput}
While goodput provides a framework for optimizing proposed length per batch, it cannot be measured directly and must be estimated before batch execution. As formulated in \autoref{eq:goodput}, calculating goodput requires estimating two key components: batch latency, which reflects current system load, and token acceptance rate, which measures the effectiveness of speculative proposals. In the following, we describe our methods for estimating these essential components.

\begin{figure}[h]
\centering
        \vspace{-0.1in}
        \includegraphics[width=0.9\linewidth]{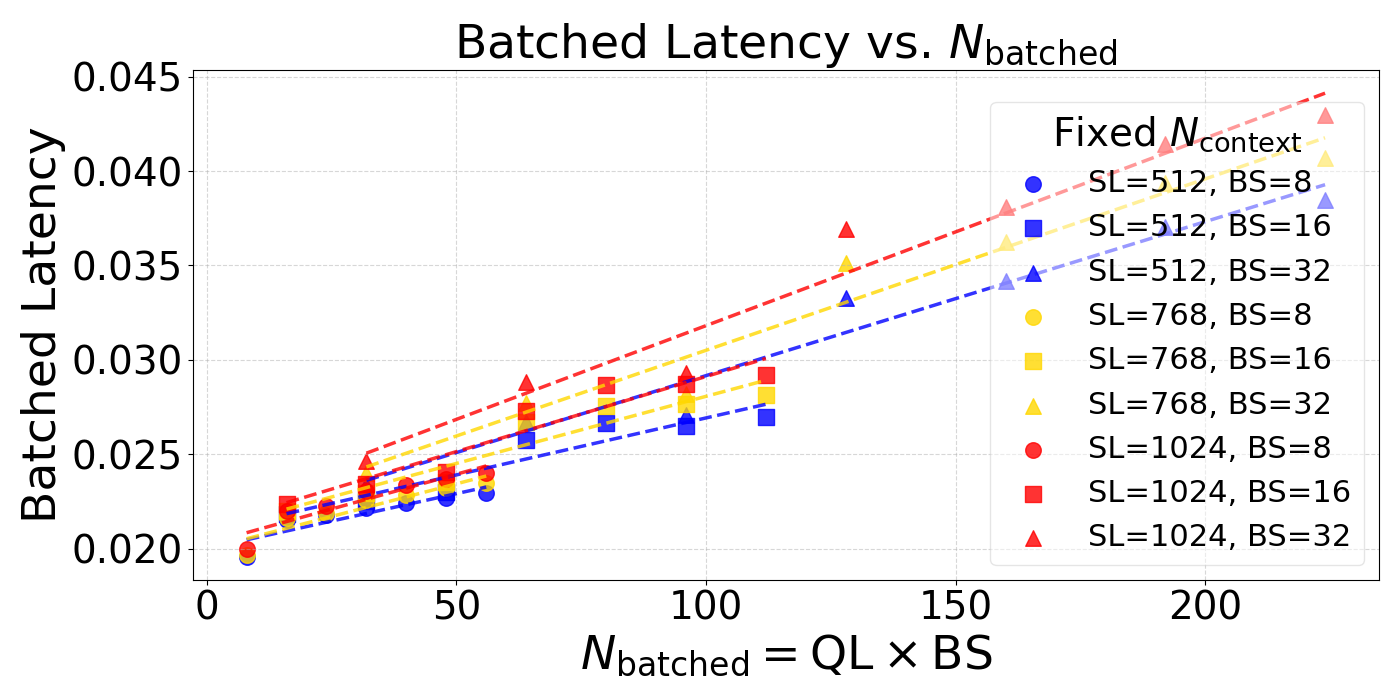}
        \vspace{-0.1in}
    \caption{Batched latency as a function of batched tokens for LLaMA3.1-70B-instruct fitted with linear regression model. QL = query length, SL = sequence length, QL = query length. Scattered dots indicate observed test set data points.}
    \label{fig:batched_latency_70b}
    \vspace{-10pt}
\end{figure}

\noindent{\textbf{Batch Latency.}} Batch latency represents the total time required to complete a speculative decoding step, encompassing both draft and target model execution times: 
\begin{equation}
    T_{batch} = T_{draft} + T_{target}
\label{eq:batch-latency}
\end{equation}
\sys employs a profiling-based method to estimate the latency. During offline profiling, it measures the target model's execution time as a function of context tokens (tokens that have already been processed and whose key-value (KV) cache is stored on the GPU) and batched tokens (tokens currently involved in the forward pass during generation). \sys fits a linear regression model to characterize the relationship between execution latency and token counts:
\begin{equation}
\begin{aligned}
    T_{fwd}(M, N_{context}, N_{batched}) &= \alpha \cdot N_{context} + \gamma \cdot N_{batched} + \delta
\end{aligned}
\label{eq:forward-time}
\end{equation}
Here, $M$ is the model and hardware configuration, $N_{context}$ is the number of context tokens within the batch and $N_{batched}$ is the number of batched tokens. The term $\alpha \cdot N_{context}$ captures KV cache loading time, $\gamma \cdot N_{batched}$ represents the computation time for scoring the proposed token in a single forward pass, and $\delta$ accounts for model-weights loading time and system-level constant overheads. Fig.~\ref{fig:batched_latency_70b} demonstrates the linear model accurately predicts batched latency for LLaMA3.1-70B-instruct across various settings.



To estimate draft method latency, \sys uses different profiling approaches depending on the speculation mechanism. Prompt lookup decoding only requires direct profiling of execution times. For draft model speculation, which generates tokens sequentially, \sys profiles forward pass latencies ($T_{fwd}$) with various batch sizes and applies the same performance model used for the target model. The total drafting time is calculated as $T_{draft} = s \times T_{fwd}$, where $s$ is the proposed length.

\noindent{\textbf{Accepted Length.}} 
\sys uses moving average to estimate token acceptance rate for a given pair of draft and target on a dataset. 
Concretely, \sys records the token acceptance rate from previous generation steps. To predict the rate in the current step, it calculates a weighted average from these past rates as~\cite{leviathan2023fast}:
\vspace{-0.1in}
\begin{equation}
\small
\label{eq:gen_len}
    l(\alpha, k) = \frac{1 - \alpha^{k+1}}{1-\alpha}
\end{equation}
Here, \(l\) is the generated length for each request including the bonus token. The variable \(\alpha\) denotes the average token acceptance rate observed within the dataset, while \(k\) corresponds to the number of tokens proposed.
We can then compute the total number of generated tokens in a single batch as 
$\sum_{i\in R} \frac{1 - \alpha_{i}^{k_i + 1}}{1 - \alpha_{i}}
\label{eq:expected-generated-tokens}
$.

\subsection{LLM Serving with \sys}
\label{sec:serving-llm-with-dsd}
\sys serving algorithm has two parts: a single generation step and a goodput optimization procedure. The generation step, shown in~\autoref{lst:decode-algo}, implements a complete pipeline for processing batches of inputs. First, \sys uses a scheduler to get the current generation batch (line 2).\footnote{We currently use the one from vLLM.} \sys only performs speculative decoding in the decoding step (lines 10-19). It begins by determining the proposed length through \texttt{GetProposedLen} (line 10), which either returns a fixed length for prompt lookup decoding or optimizes the length through \texttt{ArgMaxGoodput}(\autoref{lst:argmaxgoodput}). 
The batch then undergoes verification with length determined by \texttt{GetVerificationLen}, followed by scoring and acceptance evaluation. The process maintains a global acceptance rate that is updated after each batch processing cycle.

The goodput optimization procedure, shown in~\autoref{lst:argmaxgoodput}, implements a search strategy to find the optimal value of {\tt k} that maximizes the goodput. For each potential k value up to {\tt MAX\_LEN}, the algorithm first checks if the current proposed length will cause an out of memory error (line 5). It then estimates the acceptance length and calculates the batch latency to determine the goodput metric. 

\begin{lstlisting}[
    language=Python,
    basicstyle=\footnotesize\ttfamily,
    numbers=left,
    numberstyle=\footnotesize\color{gray},
    stepnumber=1,
    numbersep=5pt,
    backgroundcolor=\color{white},
    showspaces=false,
    showstringspaces=false,
    showtabs=false,
    frame=single,
    tabsize=4,
    captionpos=b,
    breaklines=true,
    breakatwhitespace=false,
    title=Algorithm Implementation,
    keywordstyle=\color{blue},
    commentstyle=\color{green!60!black},
    stringstyle=\color{orange},
    otherkeywords={self},
    morekeywords={def, return, if, else, for, in},
    caption={Single generation step.},
    label={lst:decode-algo},
    float=!t
]
def Generate(method):
    batch = ScheduleBatch()
    if IsPrefill(batch):
        run_proposer_prefill = True if method == 'PLD' else False
        if len(batch) > disable_batch_size:
            skip_cnt += 1
            run_proposer_prefill = False
            if skip_cnt % RESETSTEPS == 0: disable_batch_size = float('inf')
        return RunPrefill(batch, run_proposer_prefill) 
    proposed_len = GetProposedLen(method, batch)
    if proposed_len == 0:
        disable_batch_size = len(batch)
        return RunNormalDecode(batch)
    batch = Propose(batch, proposed_len)
    verification_len = GetVerificationLen(method, batch, proposed_len)
    score_results = Score(batch, verification_len)
    accept_ids, cur_acceptance_rate = Accept(score_results)
    UpdateGlobalAcceptance(cur_acceptance_rate)
    return accept_ids

def GetVerificationLen(method, batch, proposed_len):
    if method == 'PLD': return ArgMaxGoodput(batch)
    else: return proposed_len # User can plug in different pruning algorithms here
    
def GetProposedLen(method, batch):
    if method == 'PLD': return FIXED_PROPOSED_LEN
    else: return ArgMaxGoodput(batch)
\end{lstlisting}


\begin{lstlisting}[
    language=Python,
    basicstyle=\footnotesize\ttfamily,
    numbers=left,
    numberstyle=\tiny\color{gray},
    stepnumber=1,
    numbersep=5pt,
    backgroundcolor=\color{white},
    showspaces=false,
    showstringspaces=false,
    showtabs=false,
    frame=single,
    tabsize=4,
    captionpos=b,
    breaklines=true,
    breakatwhitespace=false,
    title=Algorithm Implementation,
    keywordstyle=\color{blue},
    commentstyle=\color{green!60!black},
    stringstyle=\color{orange},
    otherkeywords={self},
    morekeywords={def, return, if, else, for, in},
    caption={Finding {\tt k} to maximize goodput.},
    label={lst:argmaxgoodput},
    float=!t
]
def ArgMaxGoodput(batch):
    max_goodput, best_k = -1, 0
    for k in [1...MAX_LEN]:
        proposed_cnt = GetProposedCnt(batch)
        if OOM(proposed_cnt): continue
        accept_len = EstimateAccLen(batch, proposed_cnt)
        batch_latency = EstimateBatchLatency(batch, proposed_cnt)
        goodput = accept_len / batch_latency
        if goodput > max_goodput:
            max_goodput, best_k = goodput, k
    return best_k
\end{lstlisting}

\noindent\textbf{Prefill disabling.} 
In draft-model-based speculative decoding, the prefill phase can create computational overhead, particularly under high request rates. While speculative decoding is disabled during prefill by default, the system traditionally executes the draft model's prefill phase to maintain KV cache synchronization between draft and target models, even when no tokens are proposed in the following generation steps. This implementation can result in inefficient resource utilization.
To optimize this process, we implemented an adaptive threshold mechanism. The system records the threshold value when the proposed token length equals zero (indicating the first instance of disabled speculative decoding). For subsequent generation steps, if the batch size exceeds this threshold, the system disables the draft model's prefill phase entirely. Since this approach prevents maintenance of the draft model's KV cache for these requests, speculative decoding remains disabled for all subsequent decoding steps within these requests.
To maintain adaptability, the threshold is periodically reset, allowing the system to re-enable speculative decoding when token acceptance rates improve. 
\section{System Integration}

\begin{figure}[h]  
    \centering
    \includegraphics[width=0.9\linewidth]{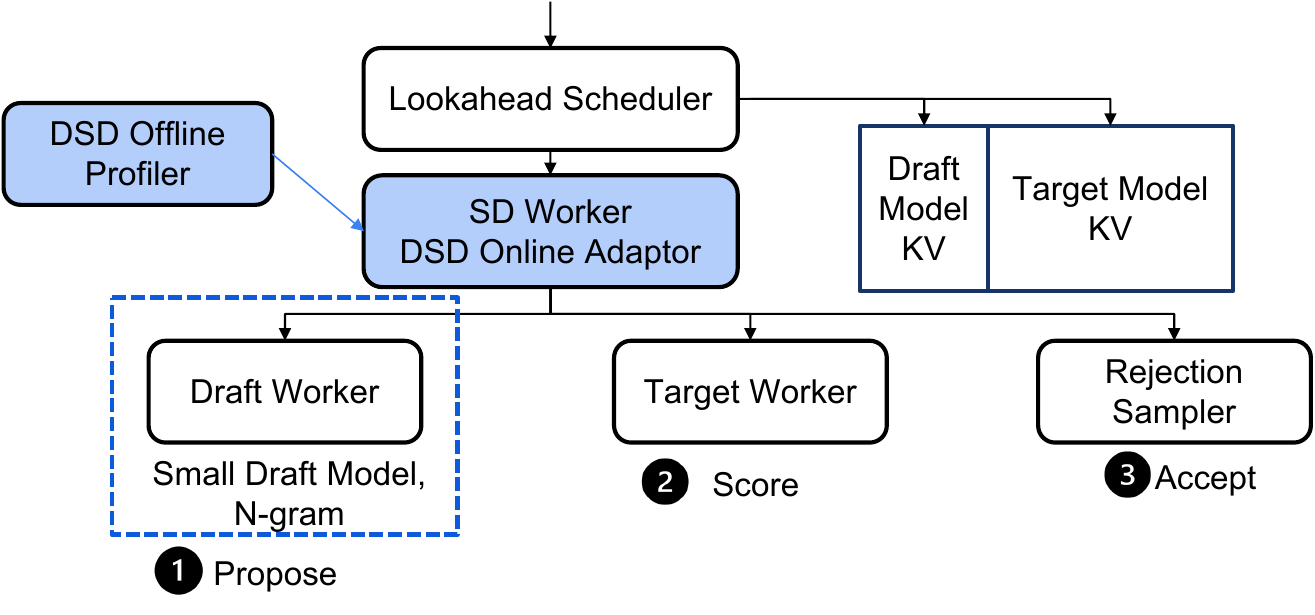}
    \caption{Architecture of \sys.}
    \label{fig:arch}
    \vspace{-0.1in}
\end{figure}

We implement \sys within vLLM~\cite{kwon2023efficient}. 
The primary modifications to vLLM for supporting \sys are twofold: (1) enabling speculative decoding and (2) dynamically determining the proposed and verification lengths.
To enable speculative decoding, we adapt the existing scheduler, memory manager, and distributed execution framework. For dynamic length determination, we introduce two additional components: an offline module and an online module, which collaborate to adjust the proposed and verification lengths based on system requirements.
\autoref{fig:arch} illustrates the system architecture. The scheduler manages active requests and allocates KV cache resources for both the draft models (if applicable) and the target models. The model runner executes the proposing, scoring, and accepting phases. Below, we detail several design choices specifically tailored for speculative decoding.

\subsection{KV Cache of Bonus Tokens}  
\begin{figure}
    \centering
    \includegraphics[width=0.99\linewidth]{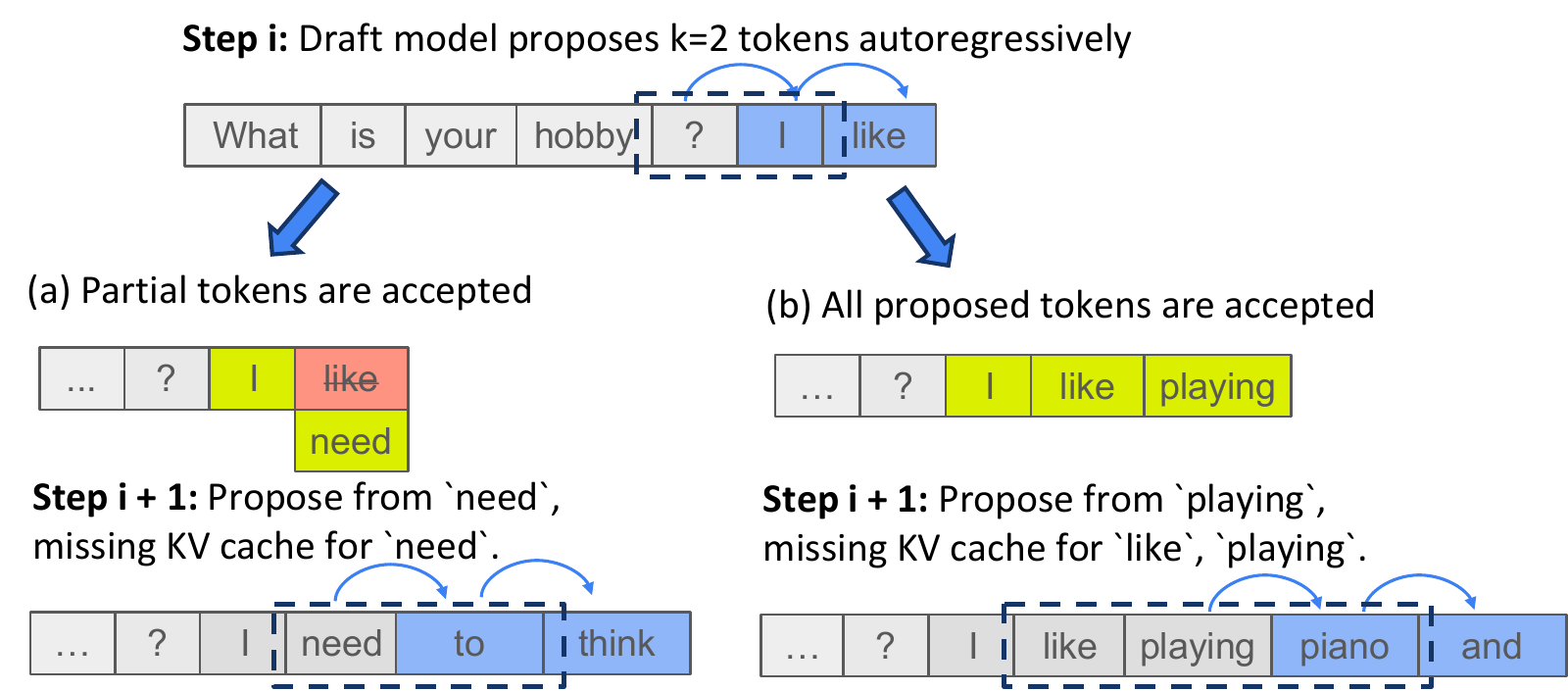}
    \caption{The draft model proposes k=2 tokens autoregressively at step $i$. (a): partial proposed tokens are accepted, step $i+1$ performs decoding and generates KV cache for `need' and `to' autoregressively. (b): all proposed tokens are accepted, step $i+1$ generates KV cache for the bonus token (` playing') and continues autoregressive generation with `playing' and `piano.' Dashed boxes indicate the tokens for which KV cache is generated in the current step.}
    \label{fig:kv-cache}
    \vspace{-13pt}
\end{figure}

The alignment KV caches between draft and target models requires precise management, particularly when handling bonus tokens. Consider a step-by-step analysis in \autoref{fig:kv-cache}:
During step $i$, the draft model operates autoregressively, generating KV cache for tokens '?' and 'I'. The KV cache for subsequent tokens ('like') remains ungenerated since they depend on the logits of previous tokens.
Two scenarios then emerge:
(1) Partial Acceptance: When only some proposed tokens are accepted, the draft model proceeds with standard decoding from the corrected token ('need').
(2) Full Acceptance: When all proposed tokens are accepted, the draft model must generate KV cache entries for both 'like' and 'playing' in step i+1.
The system maintains cache alignment through precise tracking of KV cache sizes for both models. Crucially, the draft model's forward pass must process either one or two tokens depending on the acceptance scenario. This careful management prevents cache pollution that could compromise prediction accuracy.

\subsection{Offline Profiler and Online Adaptor} 
To dynamically adjust batch sizes and prediction lengths based on real-time performance characteristics 
and token acceptance rates, \sys a two-phase adaptation strategy.
The offline profiler profiles the execution times of draft methods and target model executions across various batch sizes. Based on this profiling, it fits a model as described in \autoref{sec:estimate-goodput} to estimate execution times for different batch sizes during runtime.
The online adaptor has two primary responsibilities: (1) tracking the token acceptance rate for the current workload, and (2) leveraging the offline-derived performance model, in conjunction with the observed token acceptance rate, to dynamically determine the optimal proposed and verification lengths during online execution (follow \autoref{sec:serving-llm-with-dsd}).


\subsection{Cudagraph support}

CUDAGraph~\cite{cudagraph} is a key technique for accelerating GPU model execution, particularly during the decoding phase, which is typically memory-bound. However, effectively leveraging CUDAGraph requires developers to pre-capture the computational workload and replay it at runtime. In the decoding stage, each request generates one token per step, enabling us to bucket requests by batch size and replay them efficiently. In contrast, the prefill phase involves handling multiple requests with varying sequence lengths, making it infeasible to pre-capture all possible configurations. Fortunately, since prefill is usually compute-bound and more time-consuming, the benefits of CUDAGraph are less critical in this stage. As a result, a common practice is to apply CUDAGraph optimization only during decoding.

With speculative decoding—particularly in the verification phase—each request verifies multiple tokens, resulting in a computation pattern that resembles prefill. While this pattern is not compatible with CUDAGraph, the overhead of frequent kernel launches remains a performance concern.

To address this issue, we observe that models using continuous batching can be decomposed into alternating token-wise operations and cross-token operations. Token-wise operations are straightforward to capture with CUDAGraph. By splitting the model’s computation graph at the attention operation (the only cross-token operation), we isolate and capture only the token-wise portion using CUDAGraph. Leveraging the \texttt{torch.compile} machine learning compiler, we implement this split through a compilation pass, without modifying the model’s core execution logic. This approach accelerates the verification phase of speculative decoding.


\begin{figure*}[h!]
    \begin{minipage}{\linewidth}
        \centering
        \includegraphics[width=0.55\linewidth]{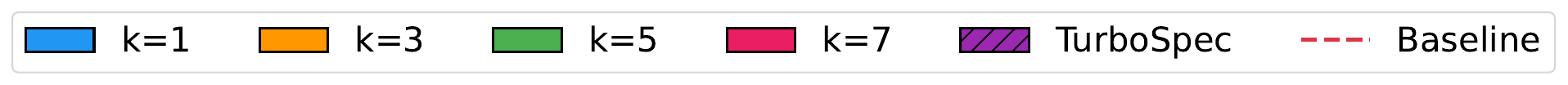}
    \end{minipage}
    \begin{subfigure}[t]{0.3\textwidth}
        \includegraphics[width=\linewidth]{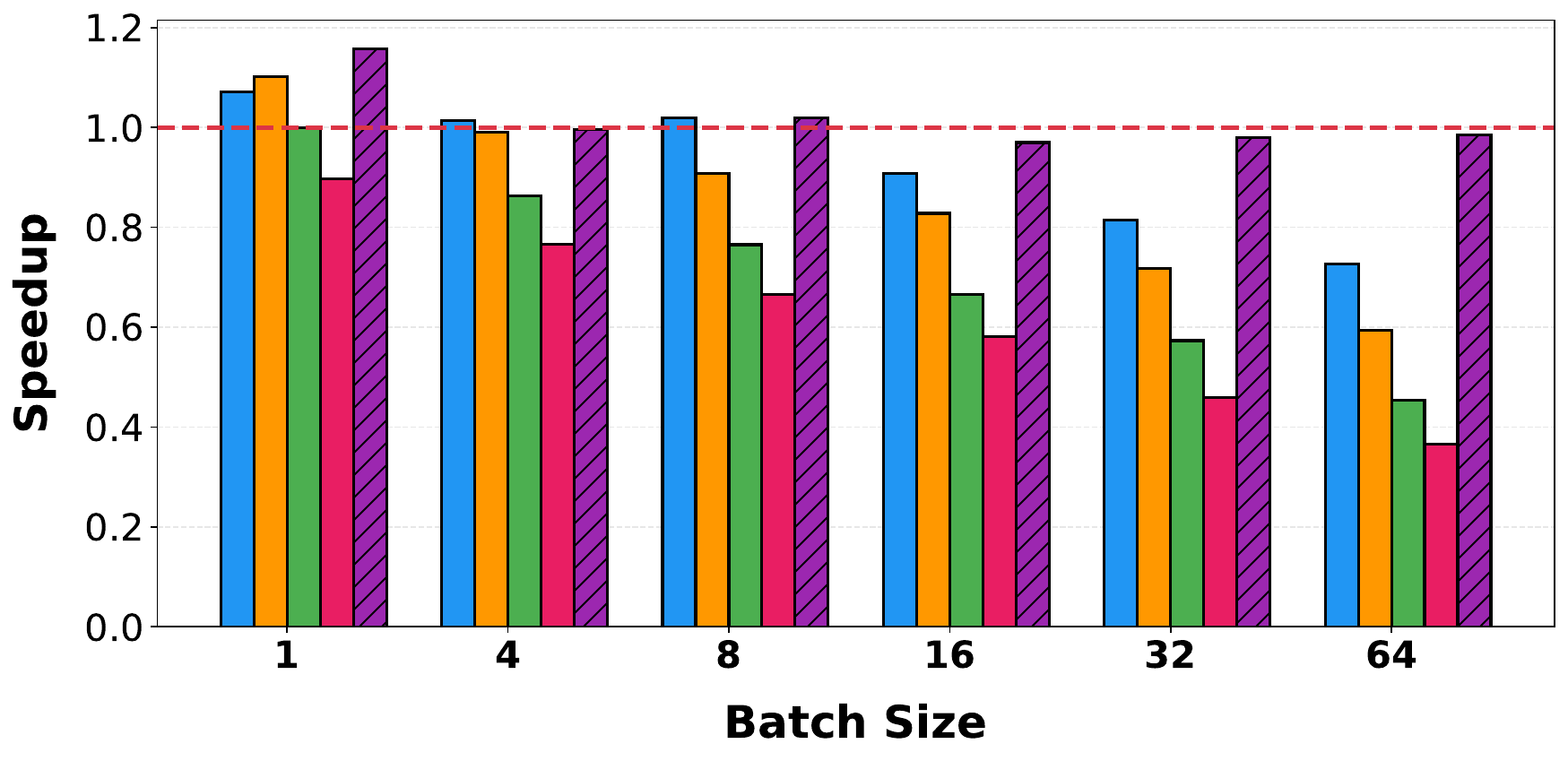}
        \caption{H100, acceptance rate = 0.5.}
        \label{fig:h100-05}
    \end{subfigure}
    \begin{subfigure}[t]{0.3\textwidth}
        \includegraphics[width=\linewidth]{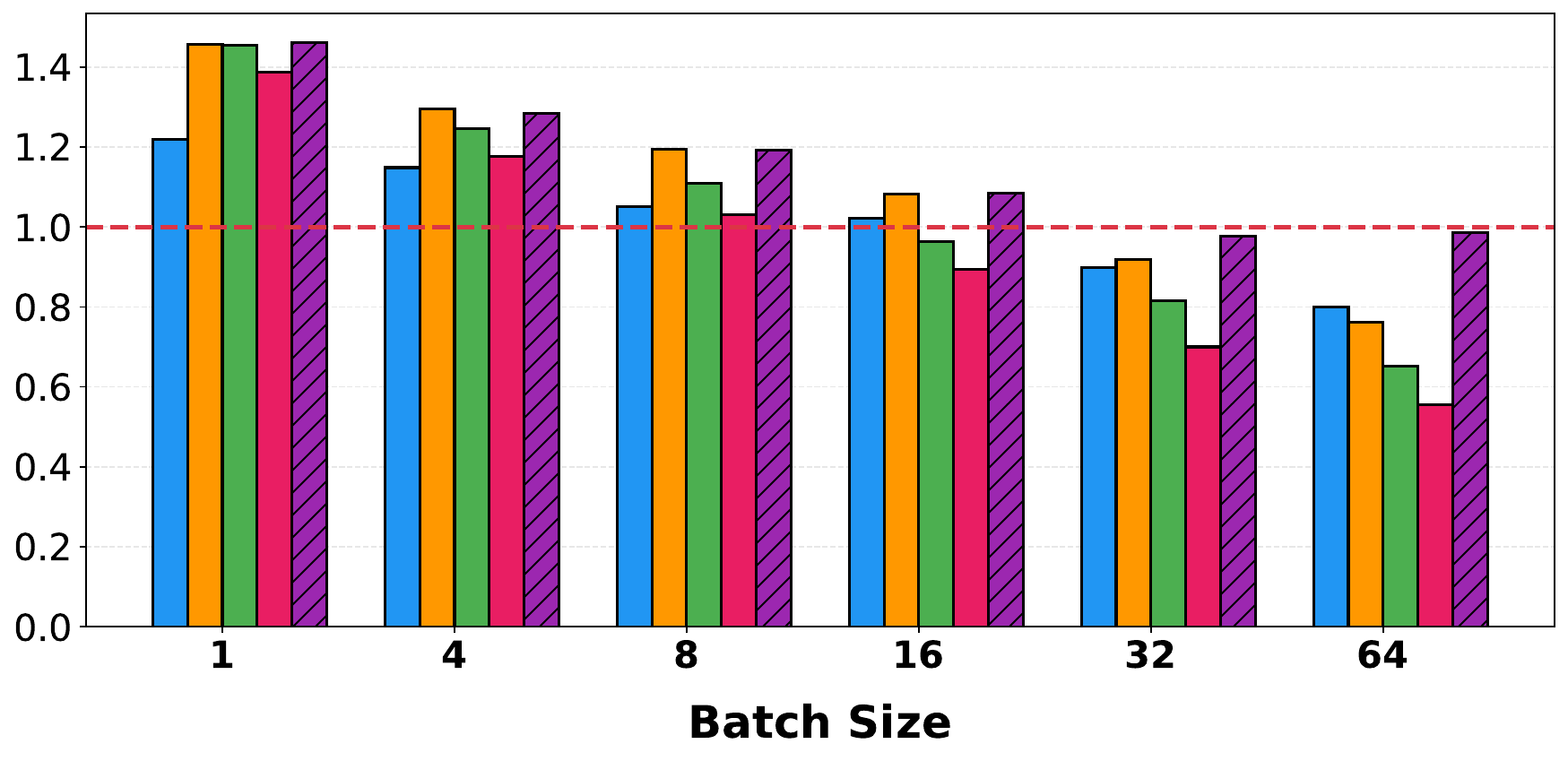}
        \caption{H100, acceptance rate = 0.7}
    \end{subfigure}
    \begin{subfigure}[t]{0.3\textwidth}
        \includegraphics[width=\linewidth]{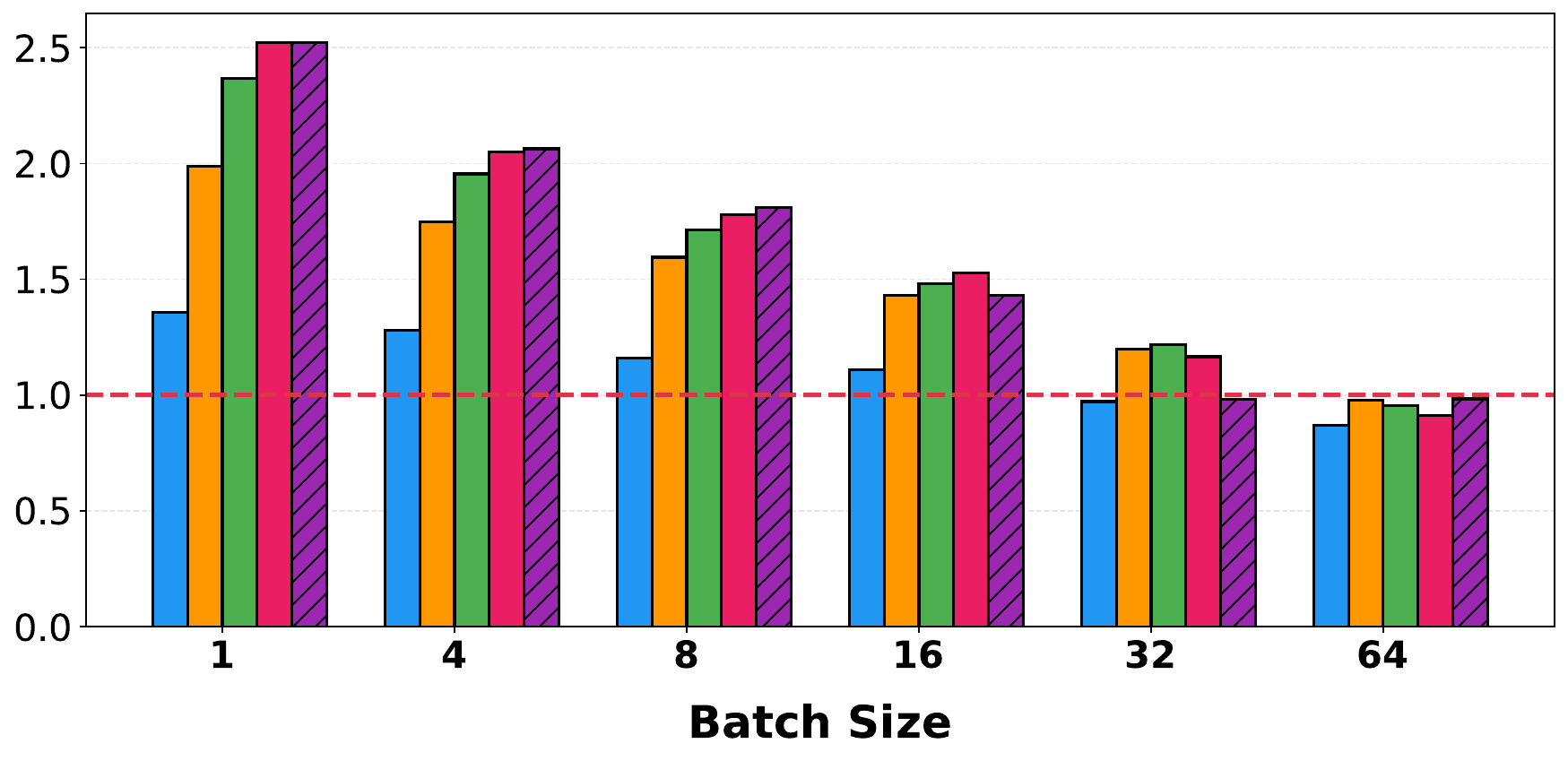}
        \caption{H100, acceptance rate = 0.9}
        \label{fig:h100-09}
    \end{subfigure}
    \begin{subfigure}[t]{0.3\textwidth}
        \includegraphics[width=\linewidth]{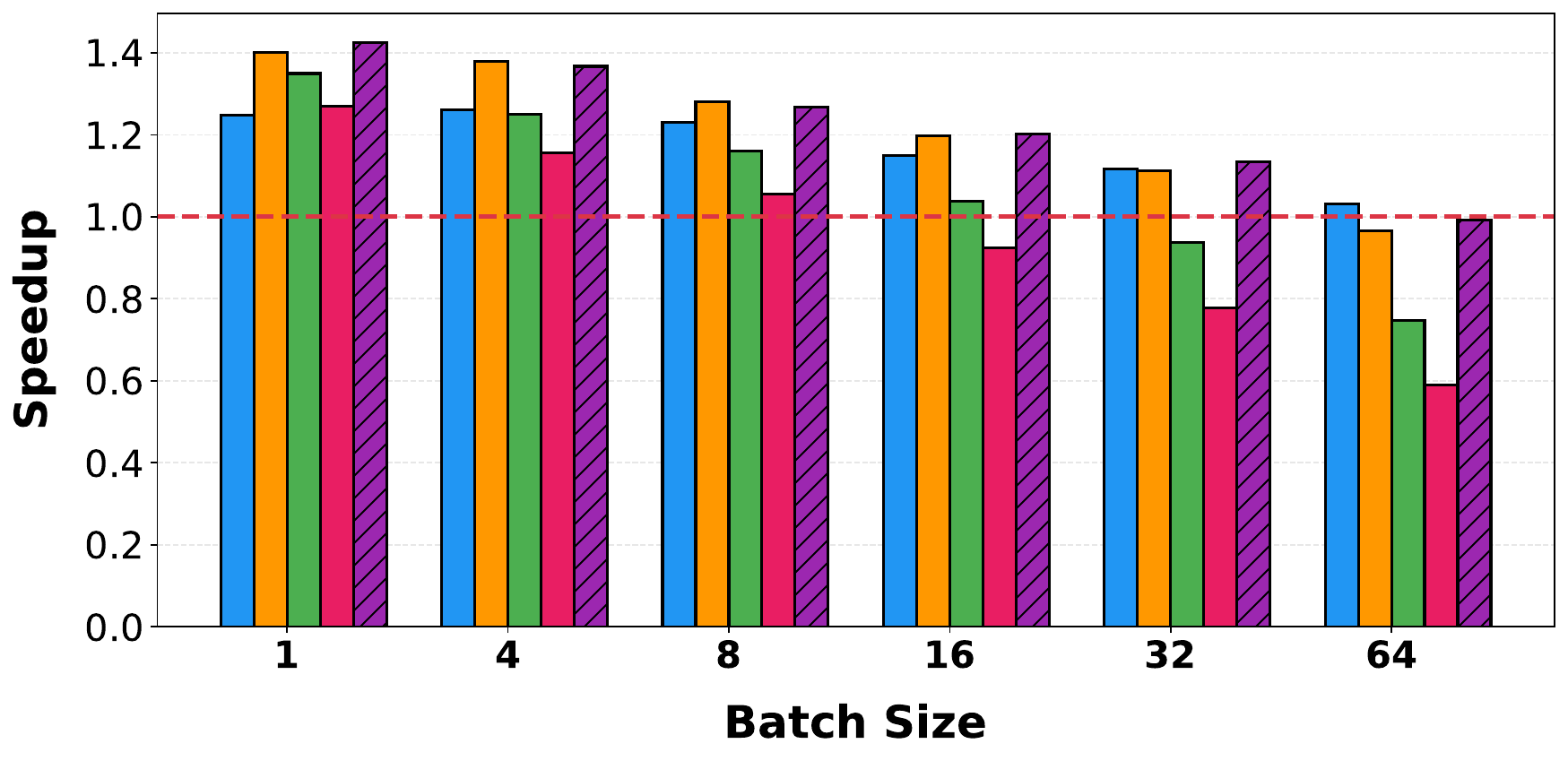}
        \caption{L40s, acceptance rate = 0.5}
    \end{subfigure}
    \begin{subfigure}[t]{0.3\textwidth}
        \includegraphics[width=\linewidth]{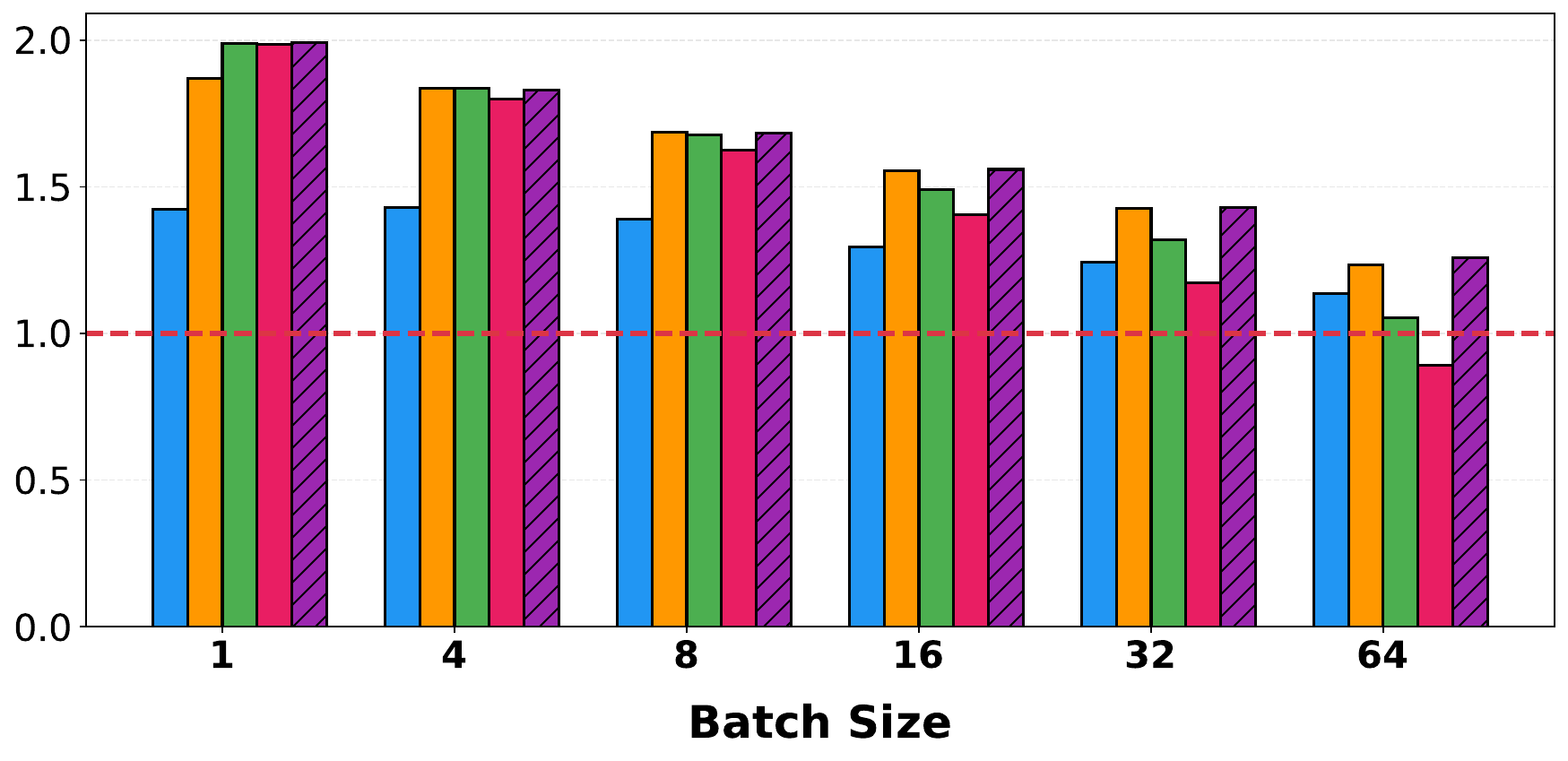}
        \caption{L40s, acceptance rate = 0.7}
    \end{subfigure}
    \begin{subfigure}[t]{0.3\textwidth}
        \includegraphics[width=\linewidth]{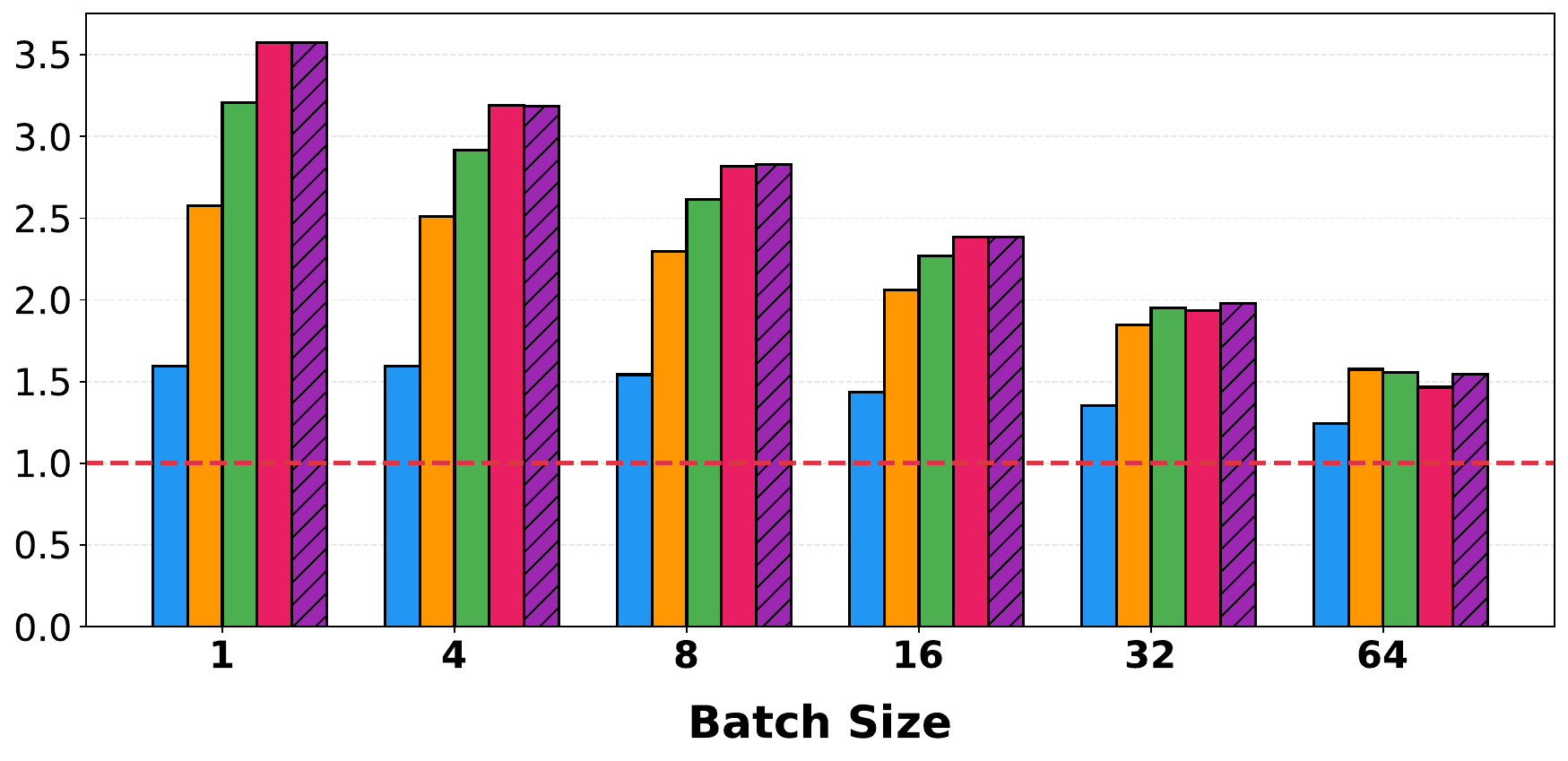}
        \caption{L40s, acceptance rate = 0.9}
    \end{subfigure}
    \caption{Speedup of \sys against static speculative decoding with fixed acceptance rate on a Llama2-7B model}
    \label{fig:fix-acc-speedup}
    \vspace{-1em}
\end{figure*}

\section{Micro Benchmark}
In this section, we evaluate \sys's efficiency under controlled token acceptance rates across hardware platforms and batch sizes in single-batch processing (\autoref{sec:micro-fixed-acc}) and offline batch inference (\autoref{sec:micro-batch-inference}).

\subsection{Request Latency for Online Serving}
\label{sec:micro-fixed-acc}
\textbf{Setup.} We evaluate Llama2-7B with fixed input/output lengths of 256 tokens NVIDIA H100 (80GB HBM)\cite{h100} and L40S (48GB HBM)\cite{l40s}.
To evaluate performance, we measure speedup as the ratio between non-speculative and speculative decoding latency per request. All experiments consist of five warmup iterations followed by ten profiling iterations, with the average request latency calculated across the profiling iterations.


\begin{figure}[h]
    \begin{center}
        \begin{minipage}{\linewidth}
            \centering
            \includegraphics[width=0.6\linewidth]{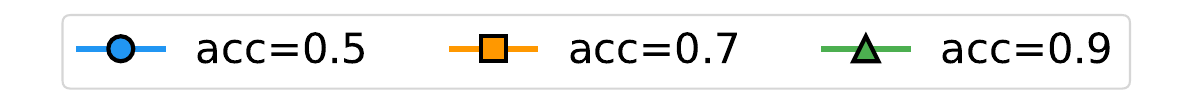}
        \end{minipage}
        \begin{subfigure}[t]{0.45\linewidth}
            \includegraphics[width=\linewidth]{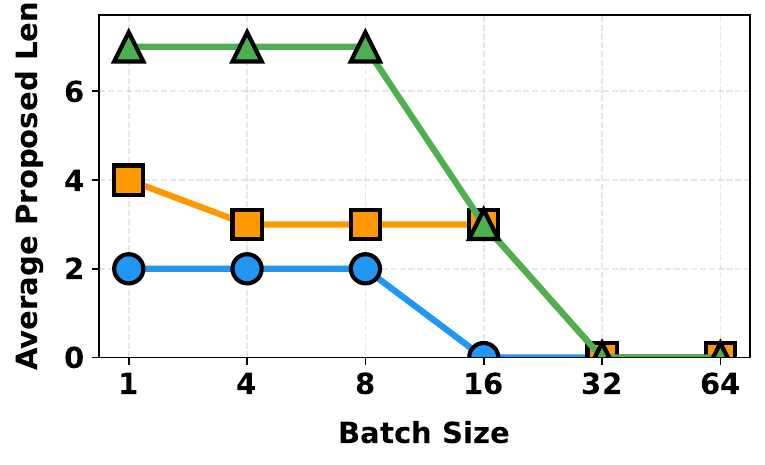}
            \caption{H100, Proposed length.}
        \end{subfigure}
        \begin{subfigure}[t]{0.45\linewidth}
            \includegraphics[width=\linewidth]{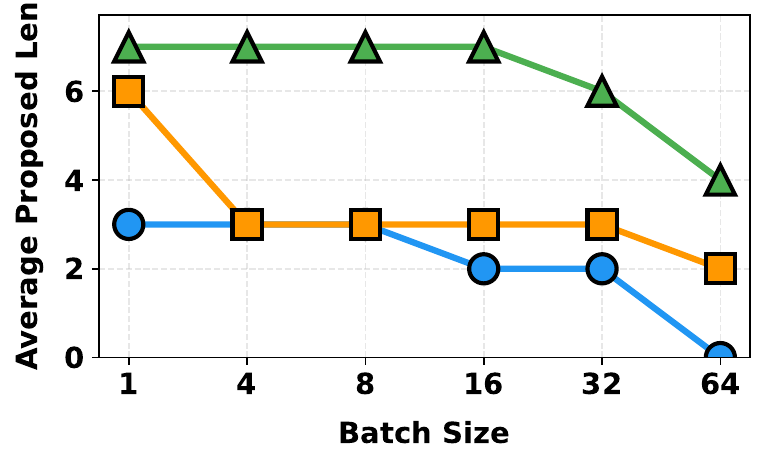}
            \caption{L40s, Proposed length.}
        \end{subfigure}
        \caption{Average proposed length of \sys across different batch sizes and hardware}
        \label{fig:proposed-len}
    \end{center}
\end{figure}

\noindent{\textbf{Results.}} As shown in \autoref{fig:fix-acc-speedup}, our experimental demonstrates that \sys achieves adaptive efficiency across three key dimensions:
(1) \sys exhibits adaptation to varying acceptance rates. As shown in \autoref{fig:proposed-len}, given the same batch size, at lower acceptance rates (0.5), the system selectively employs shorter proposed lengths (3 tokens), while at higher rates (0.9), it leverages longer proposals (7 tokens). \autoref{fig:fix-acc-speedup} empirically validates that \sys consistently determines the optimal proposal length across the entire spectrum, almost always achieving the best speedup across different acceptance rates.
(2) \sys implements dynamic deactivation of speculative decoding when computational overhead exceeds potential performance gains. This behavior is particularly evident in \autoref{fig:h100-05}, where batch sizes $\geq$16 render speculative decoding computationally inefficient. While traditional speculative decoding methods show significant performance deterioration under these conditions, \sys maintains near-baseline efficiency (speedup $\approx$0.97), with minimal overhead attributable to the speculative framework infrastructure. \autoref{fig:proposed-len} provides quantitative validation of this behavior, demonstrating that \sys automatically terminates speculation for larger batch sizes. This adaptation mechanism proves especially crucial at lower acceptance rates, where token rejection rates are substantially higher and a large amount of compute is wasted.
(3) \sys demonstrates systematic adaptation to batch size variation. As illustrated in \autoref{fig:proposed-len}, the system exhibits an inverse relationship between batch size and proposal length, optimizing computational efficiency by reducing token proposals as batch sizes increase. This dynamic adjustment ensures optimal resource utilization across computational loads.



\subsection{Throughput for Offline Batch Inference}
In this section, we investigate scenarios where speculative decoding provides advantages for offline batch inference. The fundamental principle is that speculative decoding becomes particularly effective when a system has surplus computational capacity while being constrained by memory bandwidth or capacity. We identify two specific cases where this occurs: First, on low-end GPUs with limited memory capacity, the KV cache quickly reaches its maximum capacity, resulting in restricted batch sizes even during offline processing. Second, in long-context scenarios, even high-end GPUs like the H100 with 80GB memory become constrained by memory capacity and KV cache loading requirements. In both cases, the available computational resources can be effectively utilized through speculative decoding to improve overall throughput.
\label{sec:micro-batch-inference}
\begin{figure}[h]
\begin{minipage}{\linewidth}
        \centering
        \includegraphics[width=0.9\linewidth]{figures/microbench/micro_legend.pdf}
    \end{minipage}
    \centering
    \begin{subfigure}[t]{0.48\linewidth}  
        \includegraphics[width=\linewidth]{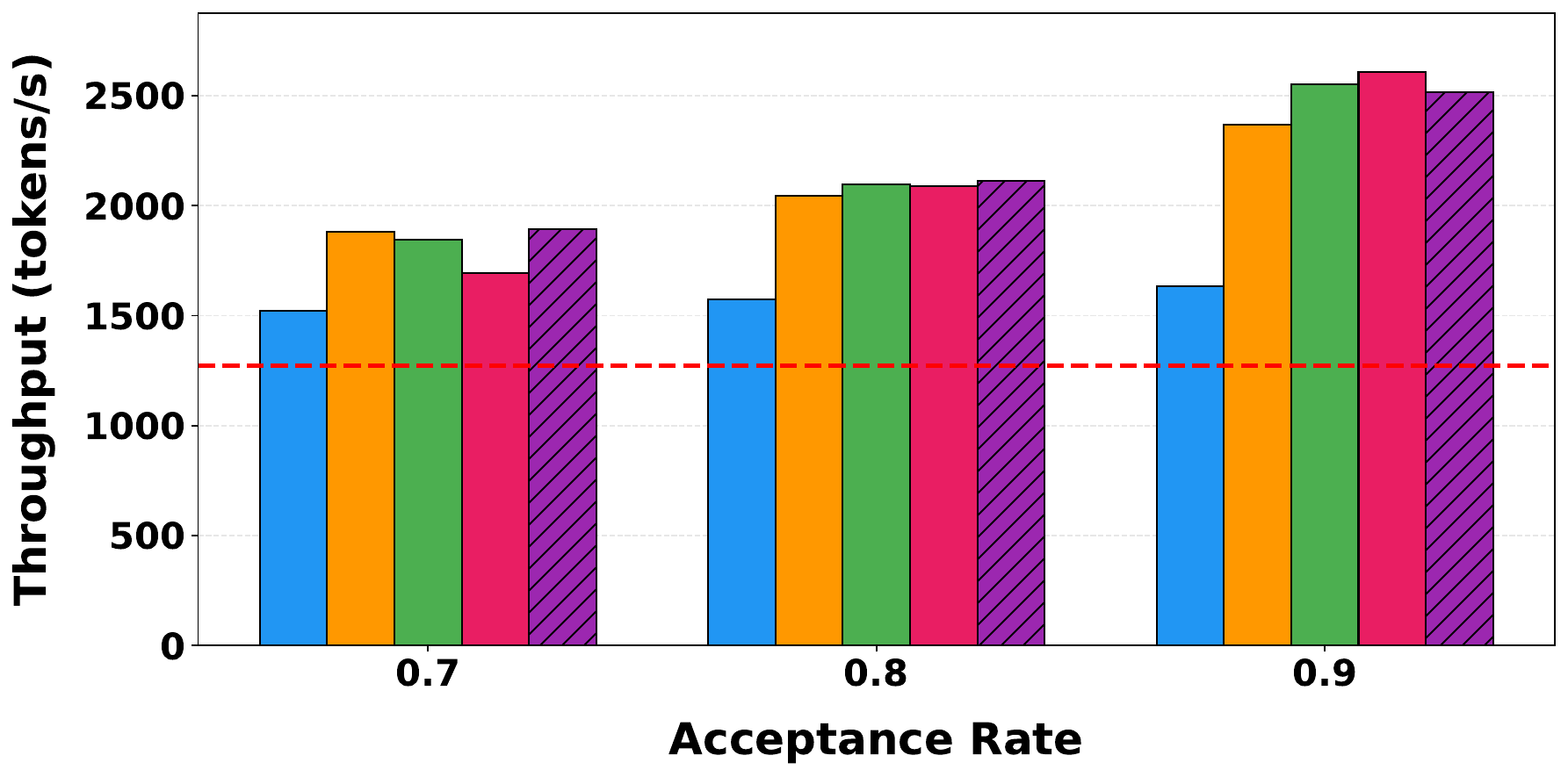}
        \caption{Low-end GPU: throughput improvement on L40S. Input length = output length = 256.}
        \label{fig:l4-tpt}
    \end{subfigure}
    \begin{subfigure}[t]{0.48\linewidth} 
        \includegraphics[width=\linewidth]{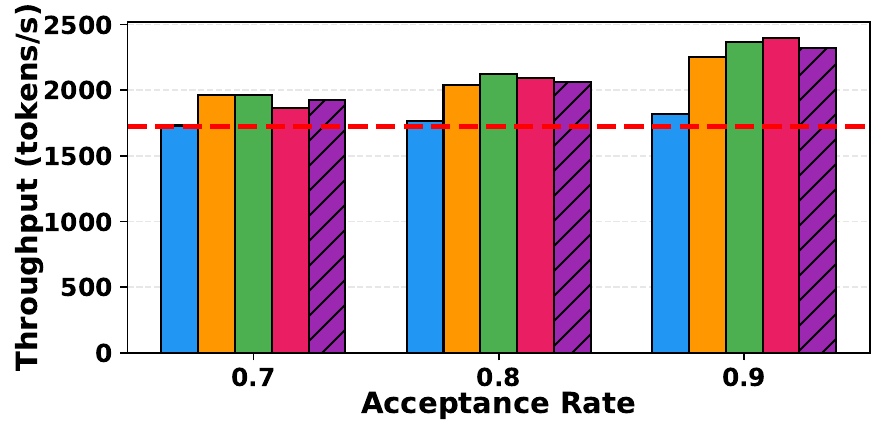}
        \caption{Long context: throughput improvement on H100. Input length = 1024, output length = 128.}
        \label{fig:h100-tpt}
    \end{subfigure}
    \vspace{-1em}
    \caption{Throughput for running LLama2-7B.}
    \vspace{-2em}
\end{figure}
We compare the throughput (the number of output tokens generated per second) with and without speculative decoding. We plot the speedup (throughput of using speculative decoding / throughput of normal decoding) in \autoref{fig:l4-tpt} and \autoref{fig:h100-tpt}.
As shown in \autoref{fig:l4-tpt}, speculative decoding provides significant throughput improvement on the L40S GPU, reaching up to nearly 2$\times$ throughput gains at higher accuracy levels (acc=0.9) compared with non-speculative decoding, with larger proposed length generally performing better. \autoref{fig:h100-tpt} reveals more modest but still notable improvements for long-context scenarios on the H100 GPU, with speedups ranging from 1.1$\times$  to 1.5$\times$  across different configurations. In both scenarios, \sys consistently demonstrates competitive or superior performance compared to fixed-length speculative approaches across different numbers of speculative tokens. This suggests its effectiveness in selecting optimal speculation lengths. 
Notably, these experiments reveal that speculative decoding extends beyond mere latency optimization. When a system has available computational resources and is not compute-bound, speculative decoding can significantly enhance offline throughput. This finding expands its utility from real-time applications to batch processing scenarios, demonstrating its versatility as a performance optimization technique.

\begin{figure*}[h]
    \begin{minipage}{\linewidth}
        \centering
        \includegraphics[width=0.35\linewidth]{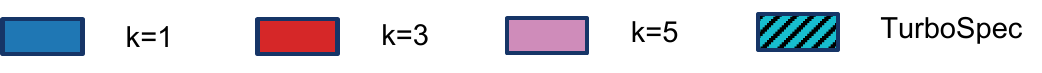}
    \end{minipage}
    \begin{subfigure}[t]{0.24\linewidth}
        \includegraphics[width=\linewidth]{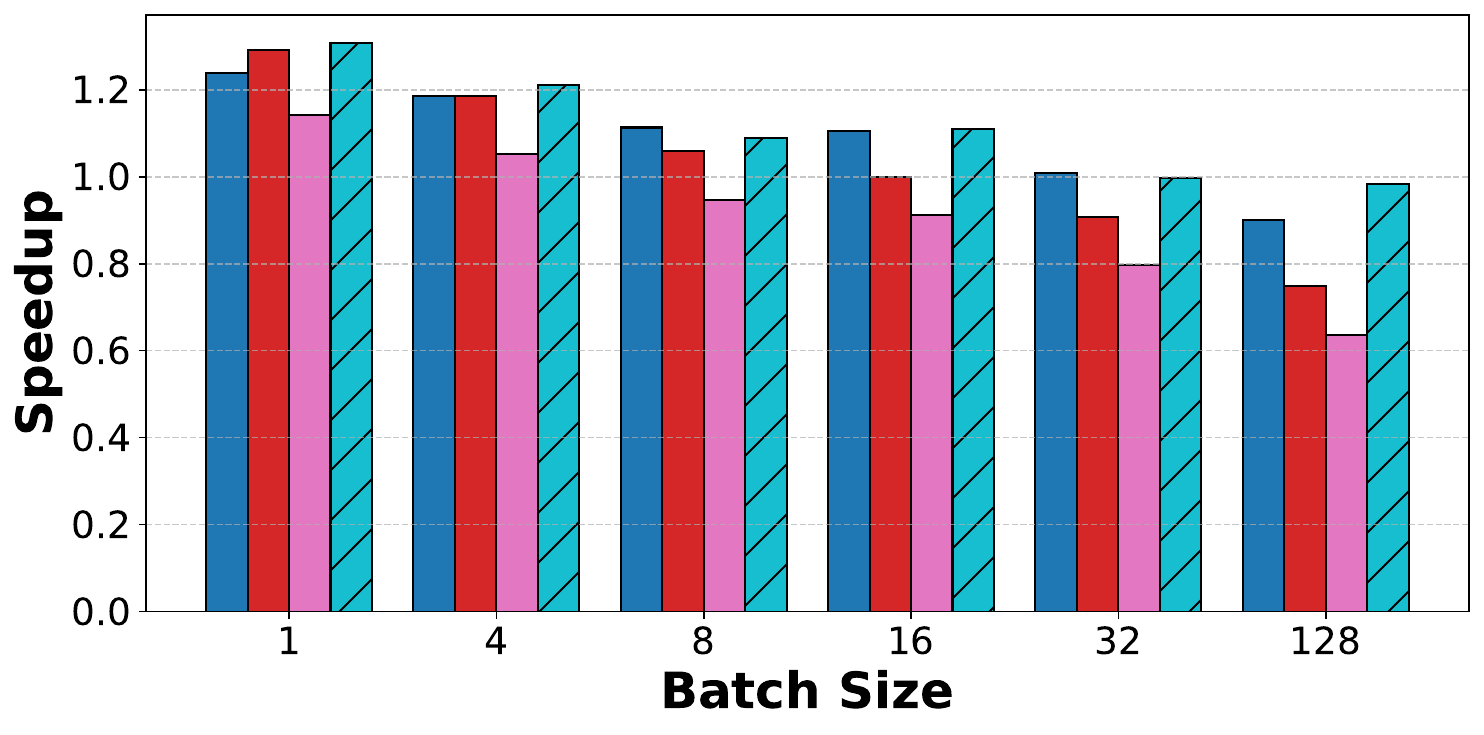}
        \caption{Eagle, 8B, ShareGPT}
        \label{fig:eagle-llama8b-sharegpt}
    \end{subfigure}
    \begin{subfigure}[t]{0.24\linewidth}
        \includegraphics[width=\linewidth]{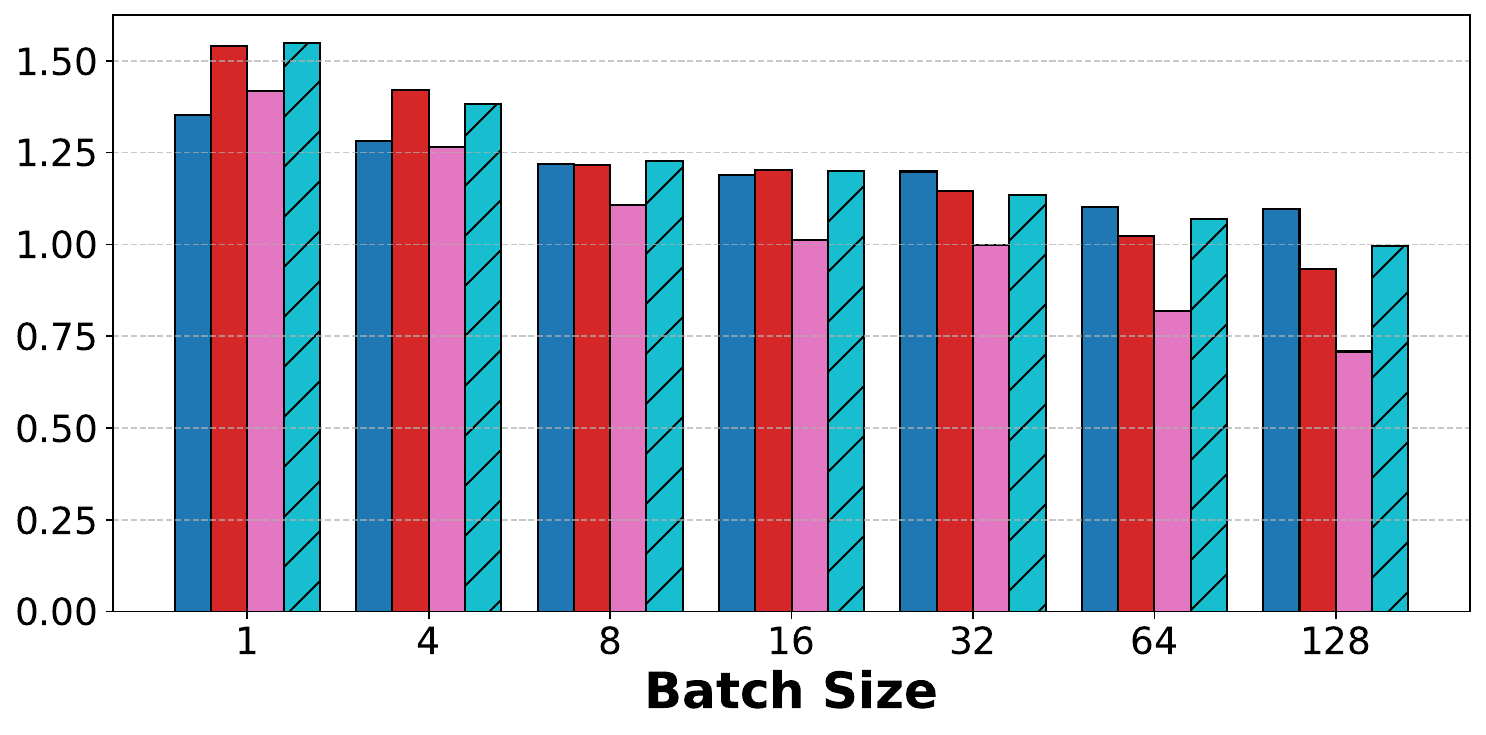}
        \caption{Eagle, 8B, InstructCode}
        \label{fig:eagle-llama8b-instructcode}
    \end{subfigure}
    \begin{subfigure}[t]{0.24\linewidth}
        \includegraphics[width=\linewidth]{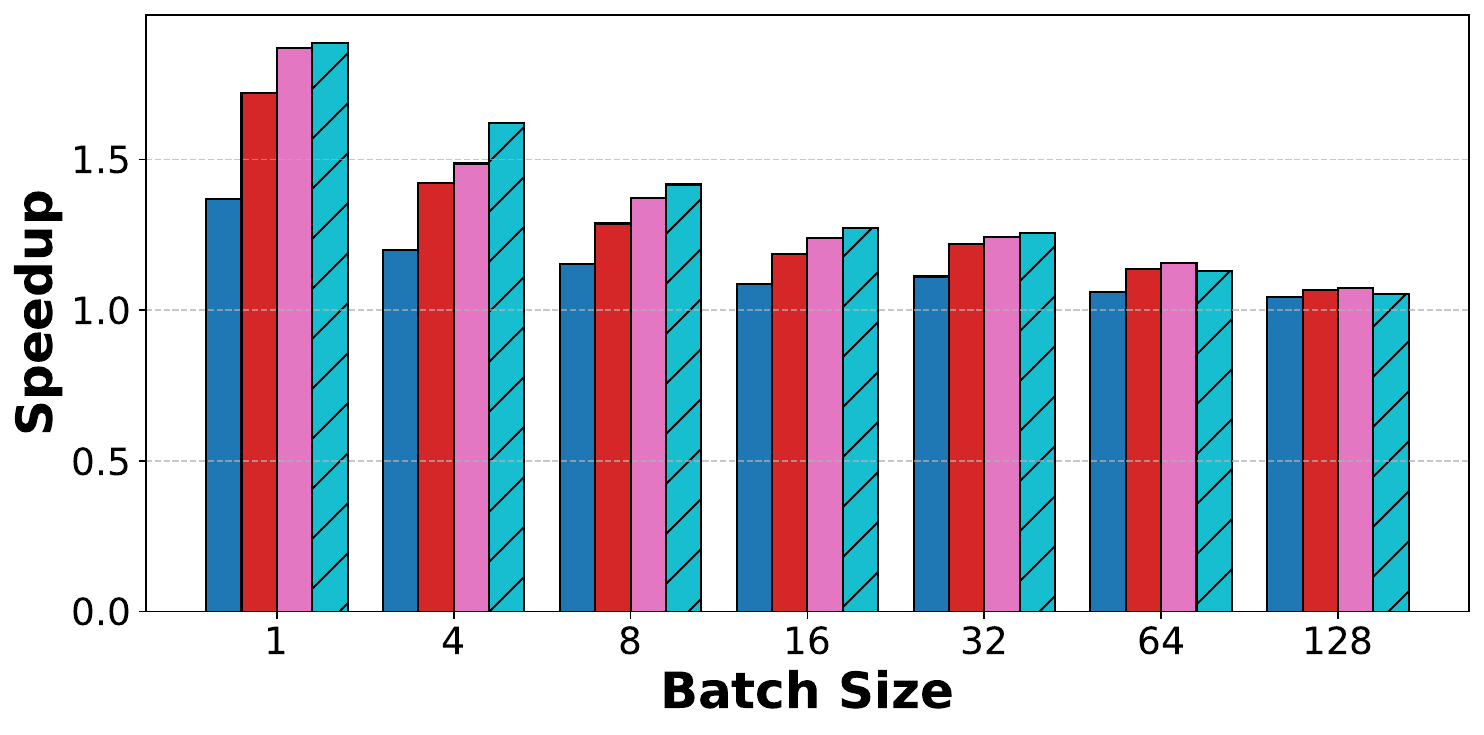}
        \caption{PLD, 8B, Sonnet.}
        \label{fig:ngram-llama8b-sonnet}
    \end{subfigure}
    \begin{subfigure}[t]{0.24\linewidth}
        \includegraphics[width=\linewidth]{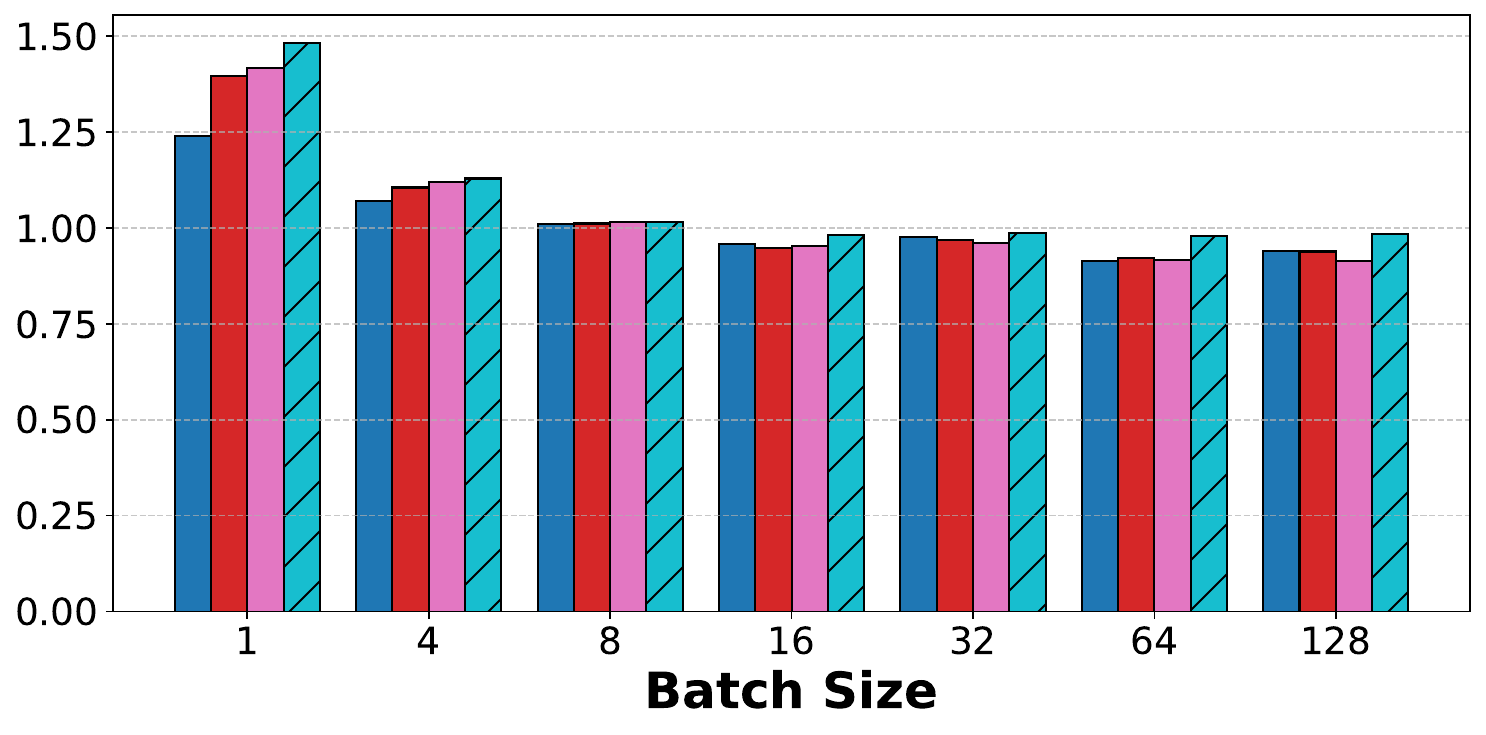}
        \caption{PLD, 8B, InstructCode}
        \label{fig:ngram-llama8b-instructcode}
    \end{subfigure}
    \begin{subfigure}[t]{0.24\linewidth}
        \includegraphics[width=\linewidth]{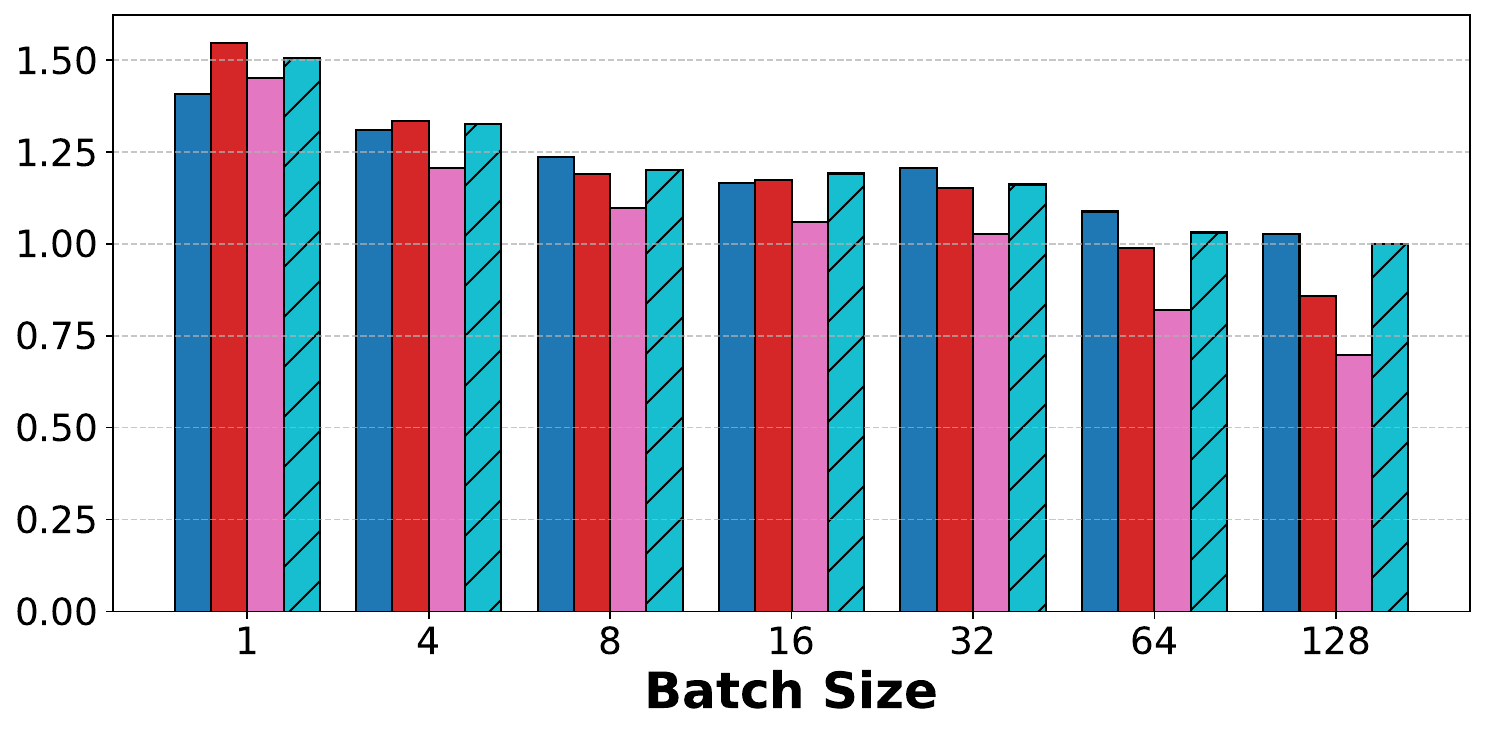}
        \caption{Eagle, 70B, ShareGPT}
        \label{fig:eagle-llama8b-sharegpt}
    \end{subfigure}
    \begin{subfigure}[t]{0.24\linewidth}
        \includegraphics[width=\linewidth]{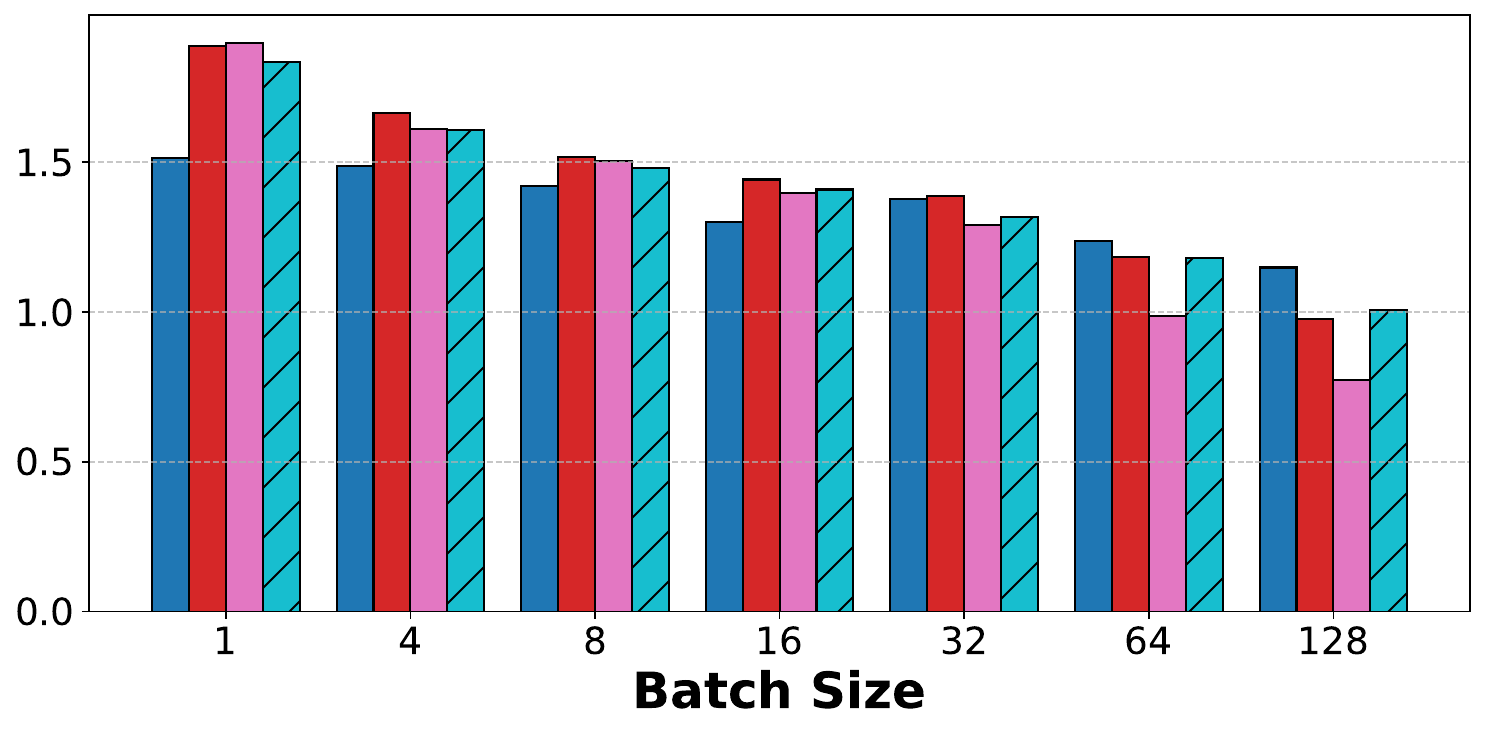}
        \caption{Eagle, 70B, InstructCode}
        \label{fig:eagle-llama70b-instructcode}
    \end{subfigure}
    \begin{subfigure}[t]{0.24\linewidth}
        \includegraphics[width=\linewidth]{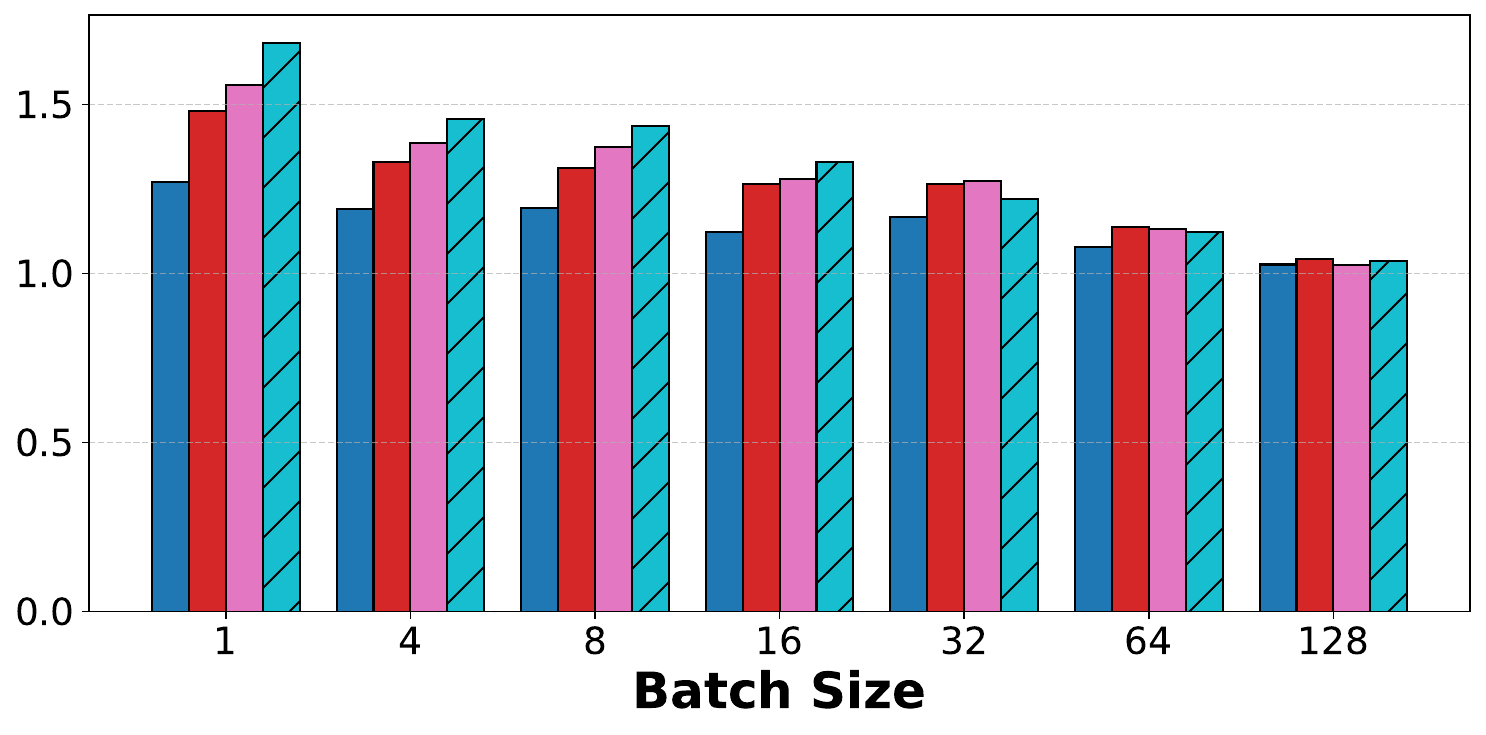}
        \caption{PLD, 70B, Sonnet.}
        \label{fig:ngram-llama70b-sonnet}
    \end{subfigure}
    \begin{subfigure}[t]{0.24\linewidth}
        \includegraphics[width=\linewidth]{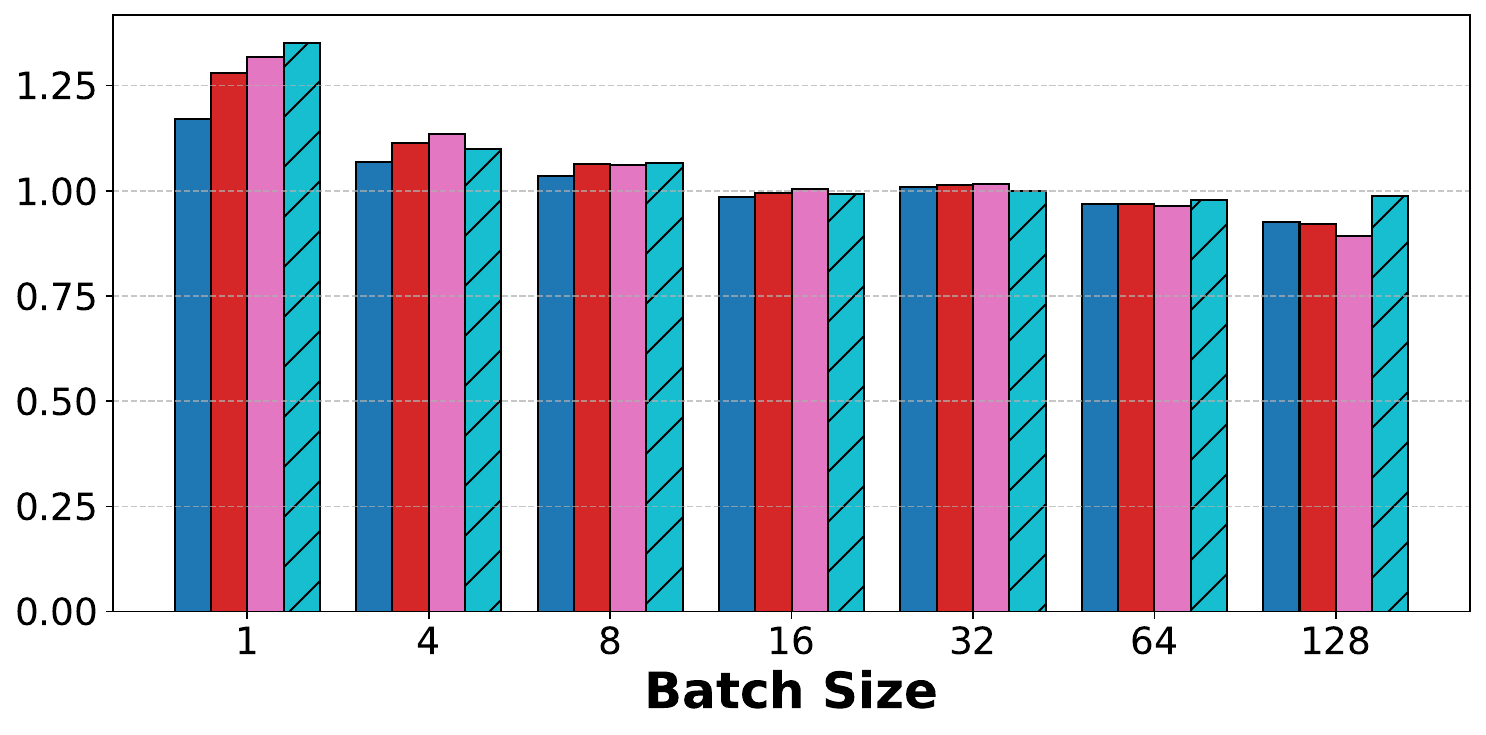}
        \caption{PLD, 70B, InstructCode}
        \label{fig:ngram-llama70b-instructcode}
    \end{subfigure}
    \vspace{-0.2in}
    \caption{Speedup on Eagle and PLD across different workloads.}
    \label{fig:speedup-eagle-ngram}
\end{figure*}

\section{End-to-end evaluation}
In this section, we evaluate the end-to-end performance of \sys on real-world datasets under both static and dynamic workload conditions.
For the static setting, we measure the average request latency across varying batch sizes using real datasets. To demonstrate the generality of \sys, we assess its effectiveness across multiple types of speculative decoding methods:
(1) Prompt lookup decoding \cite{saxena2023prompt}, where proposed tokens are retrieved as n-grams from the input prompt;
(2) Eagle \cite{li2024eagle}, where additional layers are appended to the base model to generate token predictions based on the hidden states of the final transformer layer. In our setup, each proposing head outputs the top-1 token, eliminating the need for tree attention or tree-based verification;
(3) Draft model-based decoding. Due to space constraints, results for this category are  in the supplementary materials.

We evaluate models from the LLaMA series\cite{touvron2023llama, dubey2024llama}, including LLaMA 3-8B and LLaMA 3-70B. All experiments are conducted on NVIDIA H100 GPUs with 80 GB of memory. For the 8B model, we use tensor parallelism (TP) with a degree of 1 (TP=1), and for the 70B model, we use TP=4.

\subsection{Static Workloads}
To evaluate the effectiveness of \sys, we first conduct experiments under static workloads, comparing the average request latency across different batch sizes. Specifically, we benchmark \sys against static speculative decoding (SD), where the number of speculative tokens remains fixed throughout execution. 
Note that this comparison with all possible statically proposed lengths is a strong baseline, as it involves an exhaustive sweep over all feasible speculative lengths, which in practice is impractical or infeasible due to the complexity of evaluating performance across diverse workloads and hardware configurations. Despite this, 
our results show that \sys consistently identifies optimal number of speculative tokens across all tested workloads, thus illustrating how
\sys alleviates this burden by dynamically adapting to workload characteristics in real time.
Furthermore, as shown in Figure~\ref{fig:speedup-eagle-ngram}, we observe a more significant performance degradation for Eagle as batch size increases, compared to PLD. This is attributed to the higher proposal cost in Eagle. For instance, generating three speculative tokens with Eagle on LLaMA 3-8B using an H100 GPU consumes approximately 17.7\% of the total forward pass time, whereas PLD decoding incurs only 5\% overhead for the same task.
\subsection{Dynamic Workloads}
We next evaluate \sys's adaptability in dynamic settings by varying the request rate, measured in queries per second (QPS). Specifically, we simulate a changing QPS scenario by controlling the request rate to sequentially take values of 1, 16, and 48. Each QPS level is maintained for 40 seconds, resulting in a total evaluation time of two minutes. We employ greedy decoding for all queries and measure the corresponding request latency.

\label{sec:dynamic}
\begin{figure*}[h]
\begin{minipage}{\linewidth}
        \centering
        \includegraphics[width=0.33\linewidth]{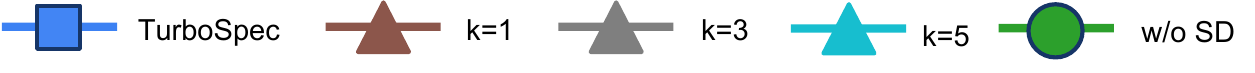}
    \end{minipage}
    \centering
    \begin{subfigure}[t]{0.24\linewidth}  
        \includegraphics[width=\linewidth]{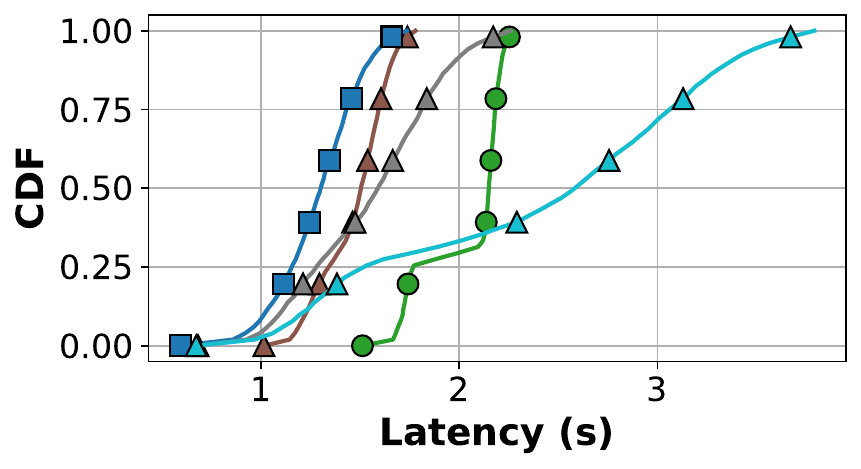}
        \caption{Eagle, InstructCode}
        \label{fig:eagle-instructcode}
    \end{subfigure}
    \begin{subfigure}[t]{0.22\linewidth} 
        \includegraphics[width=\linewidth]{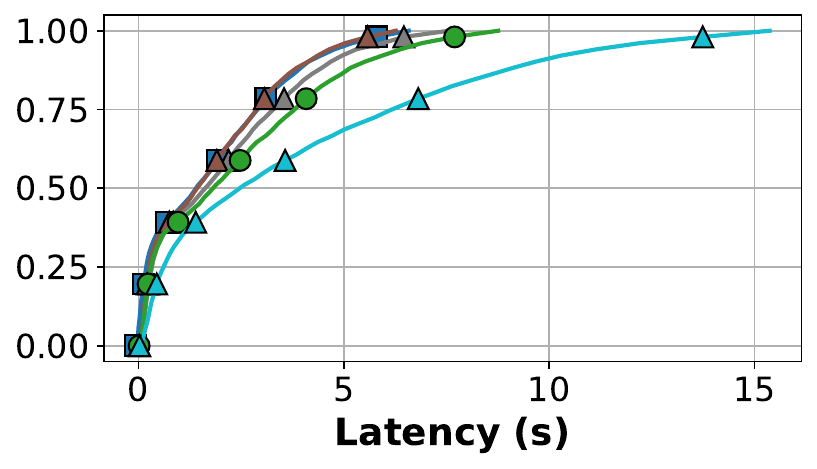}
        \caption{Eagle, ShareGPT}
        \label{fig:eagle-sharegpt}
    \end{subfigure}
    \begin{subfigure}[t]{0.22\linewidth} 
        \includegraphics[width=\linewidth]{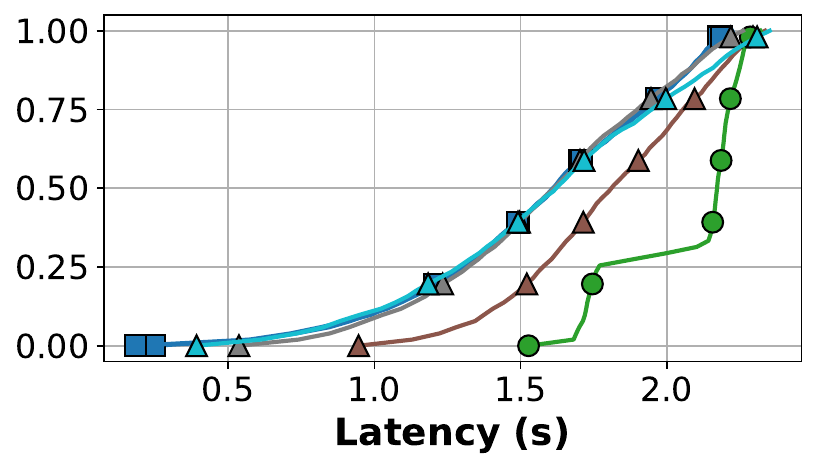}
        \caption{PLD, InstructCode}
        \label{fig:ngram-instructcode}
    \end{subfigure}
    \begin{subfigure}[t]{0.22\linewidth} 
        \includegraphics[width=\linewidth]{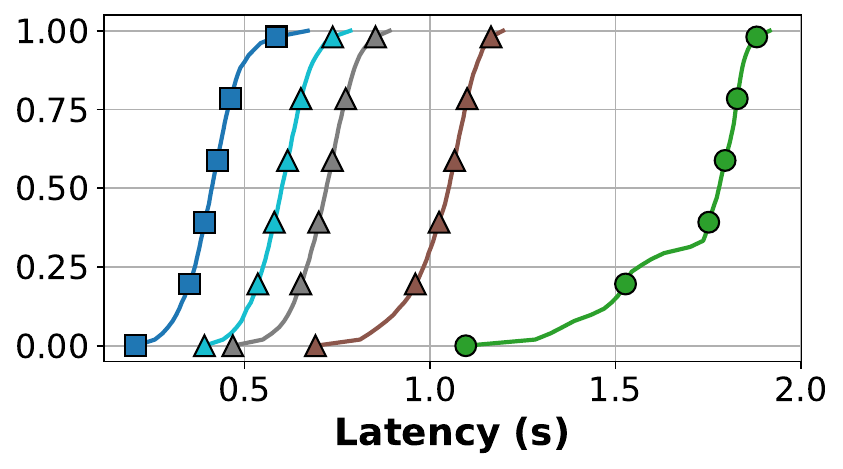}
        \caption{PLD, Sonnet}
        \label{fig:ngram-sonnet}
    \end{subfigure}
    \vspace{-0.1in}
    \caption{Request latency (s) CDF on QPS change.}
    \label{fig:dynamic_latency}
\end{figure*}

\begin{figure}[t]
    \begin{minipage}{\linewidth}
        \centering
        \includegraphics[width=0.7\linewidth]{figures/dynamic/legend-dynamic.pdf}
    \end{minipage}
    \begin{subfigure}[t]{0.5\linewidth}
        \includegraphics[width=\linewidth]{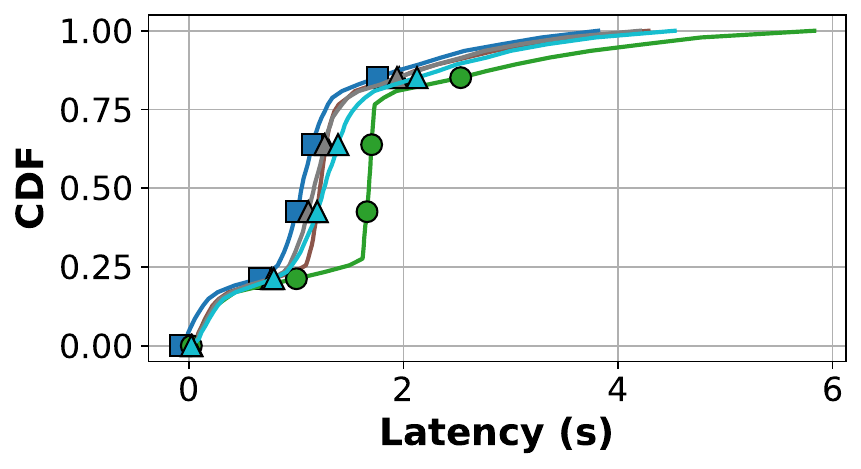}
    \end{subfigure}
    \begin{subfigure}[t]{0.48\linewidth}
        \includegraphics[width=\linewidth]{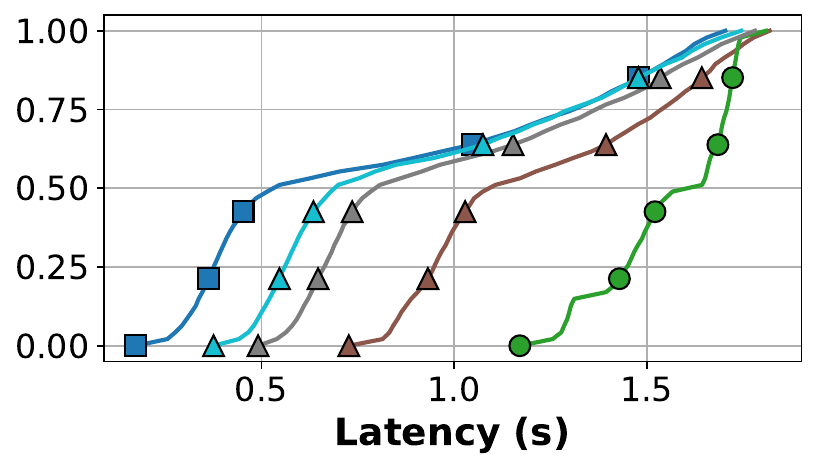}
    \end{subfigure}
    \vspace{-1em}
    \caption{Request latency (s) CDF on distribution change. The left shows the results on eagle, which shifts from InstructCode to ShareGPT. The right shows the results on PLD, which shifts from InstructCode to Sonnet.}
    \label{fig:shift}
    \vspace{-2em}
\end{figure}

\noindent{\textbf{Changing Request Rate.}}
Figure~\ref{fig:dynamic_latency} presents the CDF of request latency under dynamic QPS settings across different datasets and decoding methods. Each subplot compares the performance of \sys against several fixed configurations, including those without speculative decoding and fixed speculative decoding with k=1, k=3, and k=5.

Across all settings, \sys consistently achieves lower latencies and tighter latency distributions compared to static baselines. In the Eagle-InstructCode scenario, \sys effectively avoids the long-tail latency behavior observed in certain static configurations---particularly in cases without speculative decoding (w/o SD) and with k=5—and closely aligns with the Pareto front established by configurations k=1 and k=3. In the Eagle-ShareGPT scenario, the performance of \sys remains close to fixed speculative configurations k=1 and k=3. As illustrated in Figure~\ref{fig:eagle-sharegpt}, the latency differences between k=1 and k=3 across varying batch sizes are minimal, enabling \sys to maintain stable performance even under dynamic workloads with fluctuating batch sizes. Additionally, the notably poor performance of k=5 under high QPS conditions for both Eagle-InstructCode and Eagle-ShareGPT datasets highlights the limitations associated with overly aggressive speculative decoding under heavy workloads. Collectively, these results confirm \sys's ability to dynamically adjust speculative decoding strategies, ensuring consistently superior and stable latency performance in dynamic environments.

\noindent{\textbf{Changing Data Distribution.}}
We further evaluate \sys's adaptability under dataset distribution shifts. Specifically, we test if \sys can maintain low latency when the input distribution changes during execution. In the PLD setting (Figure~\ref{fig:shift}b), we shift the dataset from InstructCode to Sonnet, while in the Eagle setting (Figure~\ref{fig:shift}a), the dataset changes from InstructCode to ShareGPT.
In both scenarios, \sys (blue) demonstrates robust performance across the distribution shift, maintaining consistently low latency and smooth CDF curves. In contrast, static configurations exhibit clear mismatches: the no speculative decoding setting suffers from significantly higher latency, and fixed speculative lengths (k=1, k=3, k=5) fail to generalize well across the shift, often underperforming due to either insufficient or excessive speculation. These results highlight \sys's ability to adapt speculative decoding parameters in real time, even under dataset drift, ensuring stable and efficient inference performance.
\section{Ablation Study}
In this section, we perform more experiments to understand \sys's adaptability and overhead.
For all ablation experiments, we use Vicuna-160M as our draft model and Llama2-7B as the target model, on a single NVIDIA H100 GPU with 80GB of HBM3 memory. 



\begin{figure}[h]
    \centering
    \begin{subfigure}[t]{\linewidth} 
        \includegraphics[width=0.99\linewidth]{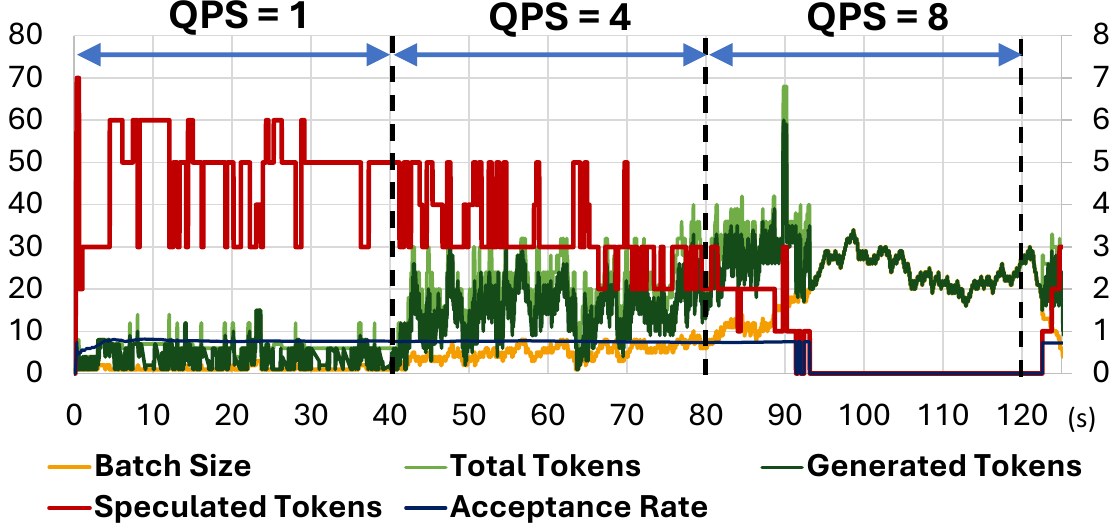}
        \caption{QPS Change}
    \label{fig:ablation_qps}
    \end{subfigure}
    \vspace{5pt}
    \begin{subfigure}[t]{\linewidth} 
        \includegraphics[width=0.99\linewidth]{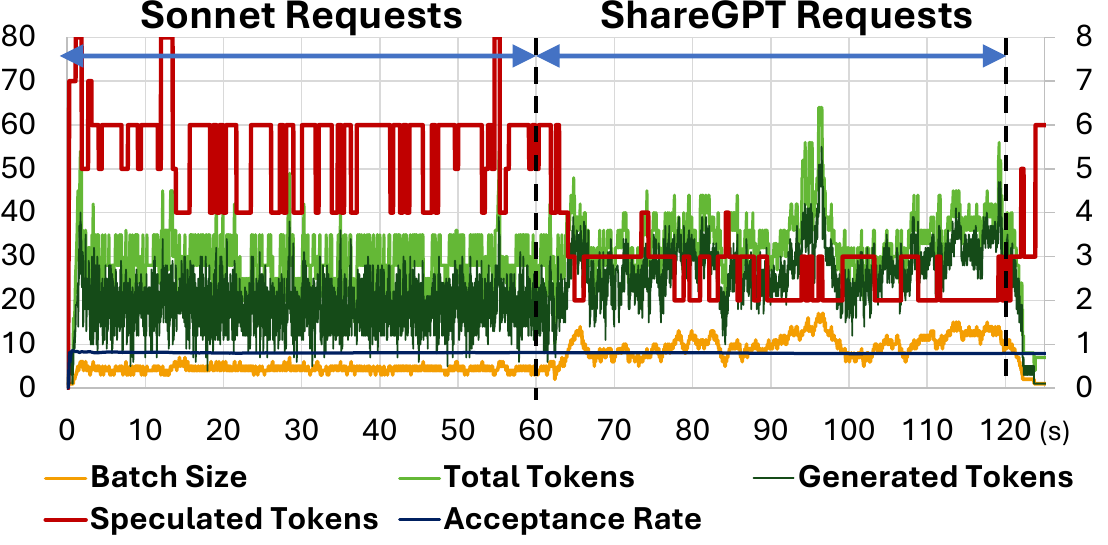}
        \caption{Dataset Change}
        \label{fig:ablation_dist}
    \end{subfigure}
    \vspace{-10pt}
    \caption{Timeline graphs of \sys operation under dynamic changes in request characteristics. Left Y axis is shared for batch size, total number of parallel tokens, and number of generated tokens, and right Y axis is shared for the number of speculated tokens and acceptance rate.}
    \vspace{-10pt}
\end{figure}

\subsection{Speculation Behavior under Dynamic Workloads}
We conducted two sets of experiments to understand \sys's adaptive behavior. First, we examined how \sys responds to varying request rates using the ShareGPT dataset. As shown in \autoref{fig:ablation_qps}, \sys dynamically adjusts its proposed speculation length based on the request rate. When the request rate reaches 8 QPS, \sys automatically disables speculative decoding to maintain system stability.
The second experiment, shown in \autoref{fig:ablation_dist}, investigates \sys's response to dataset distribution shifts. We designed a 120-second timeline where user requests transition from Sonnet to ShareGPT at the 60-second mark, maintaining a constant request rate of 6 QPS. During the Sonnet phase (0-60s), the system exhibits stable behavior with consistently high speculation rates (approximately 6 tokens per request) and relatively uniform batch sizes. However, the transition to ShareGPT requests triggers notable adaptations in system behavior: batch sizes become more variable, speculation length reduces to 2-3 tokens per request, and token throughput shows increased fluctuation. These changes reflect \sys's dynamic adaptation to ShareGPT's greater variance in input/output lengths, which directly impacts batch size variability.

\subsection{Overhead Measurement}


\begin{figure}[t]
    \begin{subfigure}[t]{0.41\linewidth}
        \includegraphics[width=0.98\linewidth]{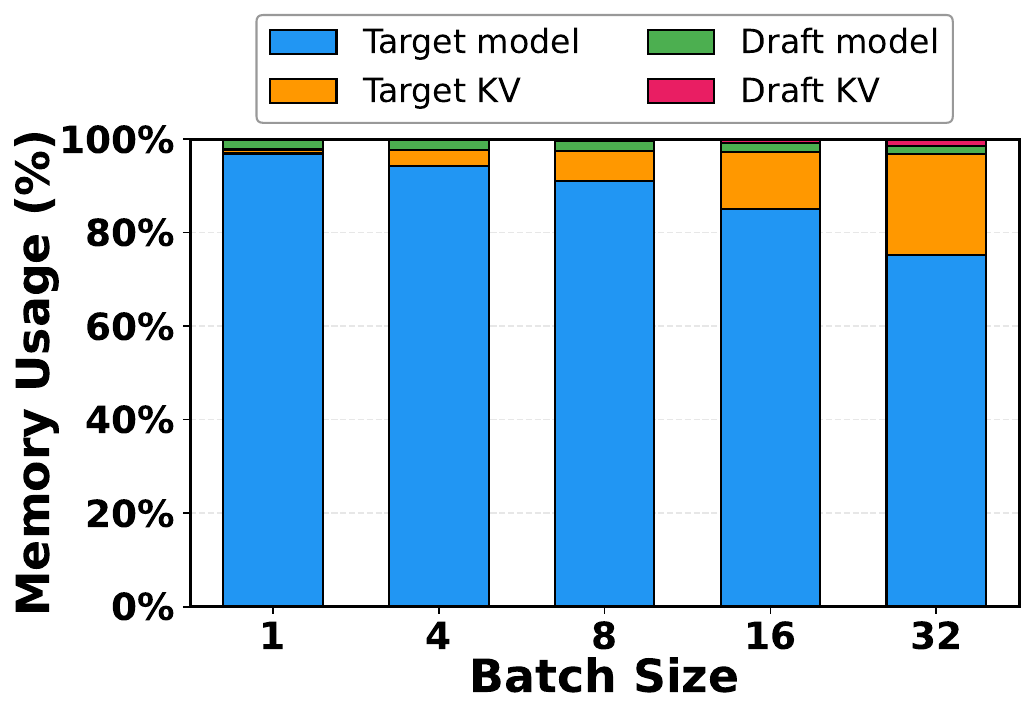}
        \caption{Memory usage}
        \label{fig:mem-breakdown}
    \end{subfigure}
    \begin{subfigure}[t]{0.58\linewidth}
        \includegraphics[width=0.98\linewidth]{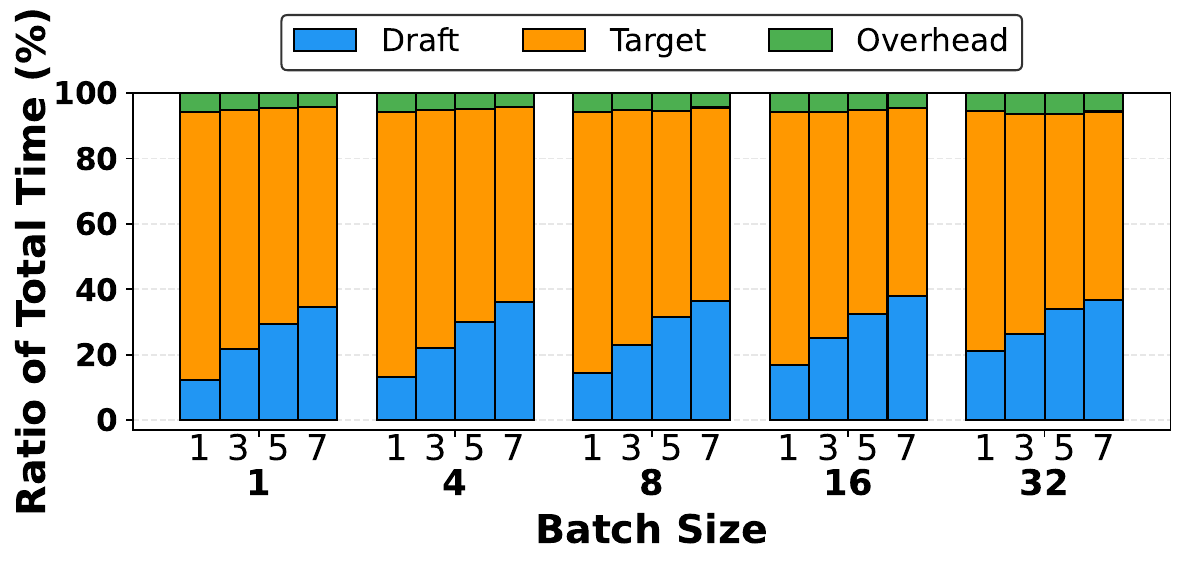}
        \caption{Execution time}
        \label{fig:70b-sharegpt}
    \end{subfigure}
    \caption{Execution breakdown of time and memory usage. We use k=1, 3, 5, 7 for the execution time breakdown.}
    \vspace{-2em}
\end{figure}

\noindent{\textbf{Memory Overhead.}} We want to point out two key characteristics of the memory footprint: (1) The proposed sequence length $k$ has no impact on the key-value (KV) cache size, as it only affects the input sequence length for a single forward pass, while the KV cache stores the cache for all previous requests. (2) Since speculative decoding performs best under small to medium batch sizes (\sys automatically disables speculative decoding for large batches where performance benefits diminish), the KV cache of the draft model and the draft model parameters constitute a minimal portion of the total memory consumption. As demonstrated in \autoref{fig:mem-breakdown}, the combined memory footprint of the draft model and its KV cache remains consistently below 4\% of total memory usage across all evaluated batch sizes, indicating negligible overhead relative to the target model's memory.

\noindent{\textbf{Execution Time Breakdown.}} The execution time breakdown reveals the relative contributions of draft model computation, target model verification, and system overhead across different batch sizes and proposal lengths. For shorter proposal lengths (k=1), the target model dominates the execution time (approximately 80\%), while the draft model accounts for only about 15\% of the total time. As the proposal length increases (k=3,5,7), the draft model's portion gradually grows, reaching around 35\% of the total time at k=7, indicating increased computational cost for longer proposals. This trend remains consistent across all batch sizes (1 to 32), though the proportions shift slightly with larger batches showing a marginally higher percentage of draft computation time. Notably, the overhead remains relatively minimal (around 5\%) across all configurations, demonstrating the efficiency of the system's implementation.

\
\vspace{-1em}
\section{Related Work}
Aside from speculative decoding (\autoref{sec:background-sd}) and continuous batching (\autoref{sec:background-sd}), the systems community has proposed many orthogonal methods to improve LLM inference performance. 

\noindent{\textbf{Quantization Methods}} such as (LLM.int8\cite{dettmers2022llmint8}, GPTQ \cite{frantar2023gptq}, Marlin \cite{frantar2024marlin}, AWQ \cite{lin2023awq}, SqueezeLLM \cite{kim2024squeezellm}) reduce the latency of LLM inference by using lower precision data types. 
Such methods trade off accuracy for performance and commonly requires a calibration step. In the context of \sys, quantization optimizations can be applied to both draft and target models to further improve our performance.

\noindent{\textbf{Prefix Caching}} techniques save compute of commonly repeated prefixes across requests.
Systems like SGLang \cite{zheng2023efficiently}, Cascade Inference \cite{cascade-inference}, and Hydragen \cite{juravsky2024hydragen} proposes efficient GPU kernels to compute and cache the computation for shared prefixes across request and deliver lower inference latency. In the context of \sys, prefix caching can be applied to both draft and target workers.

\section{Conclusion}
\sys introduces a novel approach to LLM serving by dynamically balancing inter- and intra-request parallelism through adaptive speculation management. 
Evaluation demonstrates that TurboSpec can reduce latency by up to 3.16$\times$ compared to non-speculative baselines while maintaining performance under high request rates.

\bibliographystyle{plain}
\bibliography{ref}

\begin{thebibliography}{10}

\bibitem{achiam2023gpt}
Josh Achiam, Steven Adler, Sandhini Agarwal, Lama Ahmad, Ilge Akkaya, Florencia~Leoni Aleman, Diogo Almeida, Janko Altenschmidt, Sam Altman, Shyamal Anadkat, et~al.
\newblock Gpt-4 technical report.
\newblock {\em arXiv preprint arXiv:2303.08774}, 2023.

\bibitem{sharegpt}
anon8231489123.
\newblock Sharegpt dataset, 2024.

\bibitem{sagemaker2024specdecode}
AWS.
\newblock {A}chieve up to ~2x higher throughput while reducing costs by ~50
\newblock \url{https://bit.ly/3Vyq7Rr}.
\newblock [Accessed 10-12-2024].

\bibitem{bengio2000neural}
Yoshua Bengio, R{\'e}jean Ducharme, and Pascal Vincent.
\newblock A neural probabilistic language model.
\newblock {\em Advances in neural information processing systems}, 13, 2000.

\bibitem{cai2024medusa}
Tianle Cai, Yuhong Li, Zhengyang Geng, Hongwu Peng, Jason~D Lee, Deming Chen, and Tri Dao.
\newblock Medusa: Simple llm inference acceleration framework with multiple decoding heads.
\newblock {\em arXiv preprint arXiv:2401.10774}, 2024.

\bibitem{chen2023accelerating}
Charlie Chen, Sebastian Borgeaud, Geoffrey Irving, Jean-Baptiste Lespiau, Laurent Sifre, and John Jumper.
\newblock Accelerating large language model decoding with speculative sampling.
\newblock {\em arXiv preprint arXiv:2302.01318}, 2023.

\bibitem{chen2024sequoia}
Zhuoming Chen, Avner May, Ruslan Svirschevski, Yuhsun Huang, Max Ryabinin, Zhihao Jia, and Beidi Chen.
\newblock Sequoia: Scalable, robust, and hardware-aware speculative decoding.
\newblock {\em arXiv preprint arXiv:2402.12374}, 2024.

\bibitem{vicuna2023}
Wei-Lin Chiang, Zhuohan Li, Zi~Lin, Ying Sheng, Zhanghao Wu, Hao Zhang, Lianmin Zheng, Siyuan Zhuang, Yonghao Zhuang, Joseph~E. Gonzalez, Ion Stoica, and Eric~P. Xing.
\newblock Vicuna: An open-source chatbot impressing gpt-4 with 90\%* chatgpt quality, March 2023.

\bibitem{dettmers2022llmint8}
Tim Dettmers, Mike Lewis, Younes Belkada, and Luke Zettlemoyer.
\newblock Llm.int8(): 8-bit matrix multiplication for transformers at scale, 2022.

\bibitem{dubey2024llama}
Abhimanyu Dubey, Abhinav Jauhri, Abhinav Pandey, Abhishek Kadian, Ahmad Al-Dahle, Aiesha Letman, Akhil Mathur, Alan Schelten, Amy Yang, Angela Fan, et~al.
\newblock The llama 3 herd of models.
\newblock {\em arXiv preprint arXiv:2407.21783}, 2024.

\bibitem{vicuna160m}
eqhylxx.
\newblock Vicuna-160m.
\newblock \url{https://huggingface.co/eqhylxx/vicuna-160m}, 2024.

\bibitem{frantar2024marlin}
Elias Frantar and Dan Alistarh.
\newblock Marlin: a fast 4-bit inference kernel for medium batchsizes.
\newblock \url{https://github.com/IST-DASLab/marlin}, 2024.

\bibitem{frantar2023gptq}
Elias Frantar, Saleh Ashkboos, Torsten Hoefler, and Dan Alistarh.
\newblock Gptq: Accurate post-training quantization for generative pre-trained transformers, 2023.

\bibitem{fu2024break}
Yichao Fu, Peter Bailis, Ion Stoica, and Hao Zhang.
\newblock Break the sequential dependency of llm inference using lookahead decoding.
\newblock {\em arXiv preprint arXiv:2402.02057}, 2024.

\bibitem{gao2018low}
Pin Gao, Lingfan Yu, Yongwei Wu, and Jinyang Li.
\newblock Low latency rnn inference with cellular batching.
\newblock In {\em Proceedings of the Thirteenth EuroSys Conference}, pages 1--15, 2018.

\bibitem{juravsky2024hydragen}
Jordan Juravsky, Bradley Brown, Ryan Ehrlich, Daniel~Y. Fu, Christopher Ré, and Azalia Mirhoseini.
\newblock Hydragen: High-throughput llm inference with shared prefixes, 2024.

\bibitem{sonnet2024kaggle}
Divam Kachoria.
\newblock Sonnet dataset.
\newblock \url{https://www.kaggle.com/datasets/divamkachoria/sonnetdataset}, 2024.

\bibitem{kim2024squeezellm}
Sehoon Kim, Coleman Hooper, Amir Gholami, Zhen Dong, Xiuyu Li, Sheng Shen, Michael~W. Mahoney, and Kurt Keutzer.
\newblock Squeezellm: Dense-and-sparse quantization, 2024.

\bibitem{kwon2023efficient}
Woosuk Kwon, Zhuohan Li, Siyuan Zhuang, Ying Sheng, Lianmin Zheng, Cody~Hao Yu, Joseph Gonzalez, Hao Zhang, and Ion Stoica.
\newblock Efficient memory management for large language model serving with pagedattention.
\newblock In {\em Proceedings of the 29th Symposium on Operating Systems Principles}, pages 611--626, 2023.

\bibitem{leviathan2023fast}
Yaniv Leviathan, Matan Kalman, and Yossi Matias.
\newblock Fast inference from transformers via speculative decoding.
\newblock In {\em International Conference on Machine Learning}, pages 19274--19286. PMLR, 2023.

\bibitem{li2024eagle}
Yuhui Li, Fangyun Wei, Chao Zhang, and Hongyang Zhang.
\newblock Eagle: Speculative sampling requires rethinking feature uncertainty.
\newblock {\em arXiv preprint arXiv:2401.15077}, 2024.

\bibitem{lin2024bita}
Feng Lin, Hanling Yi, Hongbin Li, Yifan Yang, Xiaotian Yu, Guangming Lu, and Rong Xiao.
\newblock Bita: Bi-directional tuning for lossless acceleration in large language models.
\newblock {\em arXiv preprint arXiv:2401.12522}, 2024.

\bibitem{lin2023awq}
Ji~Lin, Jiaming Tang, Haotian Tang, Shang Yang, Xingyu Dang, Chuang Gan, and Song Han.
\newblock Awq: Activation-aware weight quantization for llm compression and acceleration, 2023.

\bibitem{liu2023online}
Xiaoxuan Liu, Lanxiang Hu, Peter Bailis, Ion Stoica, Zhijie Deng, Alvin Cheung, and Hao Zhang.
\newblock Online speculative decoding.
\newblock {\em arXiv preprint arXiv:2310.07177}, 2023.

\bibitem{miao2023specinfer}
Xupeng Miao, Gabriele Oliaro, Zhihao Zhang, Xinhao Cheng, Zeyu Wang, Rae Ying~Yee Wong, Zhuoming Chen, Daiyaan Arfeen, Reyna Abhyankar, and Zhihao Jia.
\newblock Specinfer: Accelerating generative llm serving with speculative inference and token tree verification.
\newblock {\em arXiv preprint arXiv:2305.09781}, 2023.

\bibitem{trtllm2024specdecode}
NVIDIA.
\newblock {T}ensor{R}{T}-{L}{L}{M} {S}peculative {D}ecoding {B}oosts {I}nference {T}hroughput by up to 3.6x | {N}{V}{I}{D}{I}{A} {T}echnical {B}log --- developer.nvidia.com.
\newblock \url{https://bit.ly/41ucVAO}.
\newblock [Accessed 10-12-2024].

\bibitem{cudagraph}
NVIDIA.
\newblock Getting started with cuda graphs.
\newblock \url{https://developer.nvidia.com/blog/cuda-graphs/}, 2024.
\newblock Accessed: 2025-04-12.

\bibitem{h100}
NVIDIA.
\newblock H100.
\newblock \url{https://www.nvidia.com/en-us/data-center/h100/}, 2024.
\newblock Accessed: 2024-11-19.

\bibitem{l40s}
NVIDIA.
\newblock L40s.
\newblock \url{https://www.nvidia.com/en-us/data-center/l40s/}, 2024.
\newblock Accessed: 2024-11-19.

\bibitem{trtllm}
NVIDIA.
\newblock Tensorrt-llm.
\newblock \url{https://github.com/NVIDIA/TensorRT-LLM}, 2024.
\newblock Accessed: 2024-11-19.

\bibitem{openai2024predictedoutput}
OpenAI.
\newblock {O}pen{A}{I} {P}redicted {O}utput {R}eference.
\newblock \url{https://platform.openai.com/docs/guides/predicted-outputs}.
\newblock [Accessed 09-12-2024].

\bibitem{pope2023efficiently}
Reiner Pope, Sholto Douglas, Aakanksha Chowdhery, Jacob Devlin, James Bradbury, Jonathan Heek, Kefan Xiao, Shivani Agrawal, and Jeff Dean.
\newblock Efficiently scaling transformer inference.
\newblock {\em Proceedings of Machine Learning and Systems}, 5, 2023.

\bibitem{saxena2023prompt}
Apoorv Saxena.
\newblock Prompt lookup decoding, November 2023.

\bibitem{see-etal-2017-get}
Abigail See, Peter~J. Liu, and Christopher~D. Manning.
\newblock Get to the point: Summarization with pointer-generator networks.
\newblock In {\em Proceedings of the 55th Annual Meeting of the Association for Computational Linguistics (Volume 1: Long Papers)}, pages 1073--1083, Vancouver, Canada, July 2017. Association for Computational Linguistics.

\bibitem{somasundaram2024pld+}
Shwetha Somasundaram, Anirudh Phukan, and Apoorv Saxena.
\newblock Pld+: Accelerating llm inference by leveraging language model artifacts.
\newblock {\em arXiv preprint arXiv:2412.01447}, 2024.

\bibitem{su2023synergy}
Qidong Su, Christina Giannoula, and Gennady Pekhimenko.
\newblock The synergy of speculative decoding and batching in serving large language models.
\newblock {\em arXiv preprint arXiv:2310.18813}, 2023.

\bibitem{touvron2023llama}
Hugo Touvron, Louis Martin, Kevin Stone, Peter Albert, Amjad Almahairi, Yasmine Babaei, Nikolay Bashlykov, Soumya Batra, Prajjwal Bhargava, Shruti Bhosale, et~al.
\newblock Llama 2: Open foundation and fine-tuned chat models.
\newblock {\em arXiv preprint arXiv:2307.09288}, 2023.

\bibitem{vllm2024serving}
{vLLM Contributors}.
\newblock {vLLM Benchmark Serving}, 2024.

\bibitem{wang2024minions}
Siqi Wang, Hailong Yang, Xuezhu Wang, Tongxuan Liu, Pengbo Wang, Xuning Liang, Kejie Ma, Tianyu Feng, Xin You, Yongjun Bao, et~al.
\newblock Minions: Accelerating large language model inference with adaptive and collective speculative decoding.
\newblock {\em arXiv preprint arXiv:2402.15678}, 2024.

\bibitem{wang2024burstgpt}
Yuxin Wang, Yuhan Chen, Zeyu Li, Xueze Kang, Zhenheng Tang, Xin He, Rui Guo, Xin Wang, Qiang Wang, Amelie~Chi Zhou, and Xiaowen Chu.
\newblock Burstgpt: A real-world workload dataset to optimize llm serving systems, 2024.

\bibitem{cascade-inference}
Zihao Ye, Ruihang Lai, Bo-Ru Lu, Chien-Yu Lin, Size Zheng, Lequn Chen, Tianqi Chen, and Luis Ceze.
\newblock Cascade inference: Memory bandwidth efficient shared prefix batch decoding, February 2024.

\bibitem{yu2022orca}
Gyeong-In Yu, Joo~Seong Jeong, Geon-Woo Kim, Soojeong Kim, and Byung-Gon Chun.
\newblock Orca: A distributed serving system for $\{$Transformer-Based$\}$ generative models.
\newblock In {\em 16th USENIX Symposium on Operating Systems Design and Implementation (OSDI 22)}, pages 521--538, 2022.

\bibitem{zheng2023efficiently}
Lianmin Zheng, Liangsheng Yin, Zhiqiang Xie, Jeff Huang, Chuyue Sun, Cody~Hao Yu, Shiyi Cao, Christos Kozyrakis, Ion Stoica, Joseph~E. Gonzalez, Clark Barrett, and Ying Sheng.
\newblock Efficiently programming large language models using sglang, 2023.

\bibitem{zheng2024sglang}
Lianmin Zheng, Liangsheng Yin, Zhiqiang Xie, Chuyue Sun, Jeff Huang, Cody~Hao Yu, Shiyi Cao, Christos Kozyrakis, Ion Stoica, Joseph~E Gonzalez, et~al.
\newblock Sglang: Efficient execution of structured language model programs.
\newblock {\em arXiv preprint arXiv:2312.07104}, 2024.

\bibitem{zhou2023distillspec}
Yongchao Zhou, Kaifeng Lyu, Ankit~Singh Rawat, Aditya~Krishna Menon, Afshin Rostamizadeh, Sanjiv Kumar, Jean-Fran{\c{c}}ois Kagy, and Rishabh Agarwal.
\newblock Distillspec: Improving speculative decoding via knowledge distillation.
\newblock {\em arXiv preprint arXiv:2310.08461}, 2023.

\end{thebibliography}

\appendix
\clearpage
\section{Draft-model Based Speculative Decoding Results}
\label{sec:eval-e2e}
\begin{table}[t]
    \centering
    \caption{Hardware and model configurations. Memory in GB. TP is the tensor parallel size of the given model.}
    \label{tab:hw-config}
    \resizebox{0.99\linewidth}{!}{ 
    \begin{tabular}{llll}
        \toprule
        Hardware & Memory & Draft Model (TP) & Target Model (TP) \\
        \midrule
        L40S~\cite{l40s}             & 48            & Llama-160M~\cite{miao2023specinfer} (1) & Vicuna-7B~\cite{vicuna2023}  (1) \\
        L40S~\cite{l40s}             & 48            & Llama-160M~\cite{miao2023specinfer}  (1) & Vicuna-13B~\cite{vicuna2023}  (1) \\
        2 $\times$ H100~\cite{h100} & 2 $\times$ 80 & Llama-160M~\cite{miao2023specinfer}  (1) & Vicuna-33B~\cite{vicuna2023}  (2) \\
        4 $\times$ H100~\cite{h100} & 4 $\times$ 80 & Llama3.2-1B~\cite{dubey2024llama}  (1) & Llama3.1-70B~\cite{dubey2024llama}  (4) \\
        \bottomrule
    \end{tabular}
    }
    \vspace{-10pt}
\end{table}


\noindent{\textbf{Model and server configurations.}}
For draft model-based methods, we employ Llama-160M as the draft model ~\footnote{We fine-tune the original LLama-160M model~\cite{miao2023specinfer} on ShareGPT dataset to improve token acceptance rate~\cite{vicuna160m}} and various target models ranging from Vicuna-7B to Llama3.1-70B. For prompt lookup-based methods, we use the same target models without a draft model. The hardware setup consists of single NVIDIA L40S GPUs (48GB memory) for smaller models, scaling to two H100s (160GB total) or four H100s (320GB total) for larger models. In draft model-based configurations, the draft model runs with tensor parallelism size one (TP=1) to minimize communication overhead, while target models employ tensor parallelism up to 4-way as required by model size. For prompt lookup decoding, the target models have the same tensor parallelism settings as the draft model based method and we use CPU for proposing tokens.

\noindent{\textbf{Workloads and methods.}}
We evaluate our system across three distinct workload categories, each with unique characteristics. For online chatting, we use the ShareGPT dataset~\cite{sharegpt}, which features medium-length inputs and longer outputs with high variance. The writing completion workload utilizes the Sonnet dataset~\cite{sonnet2024kaggle, vllm2024serving}, while text summarization employs the CNN/Daily Mail dataset~\cite{see-etal-2017-get}, characterized by long inputs and consistently shorter outputs. We implement draft-model based speculative decoding for both ShareGPT and Sonnet datasets, while the CNN/Daily Mail workload uses prompt lookup style speculative decoding. Detailed workload statistics, including average input and output lengths, are presented in \autoref{tab:dataset-stats-1} and \autoref{tab:dataset-stats-2}. For comprehensive evaluation, we simulate request patterns using a Poisson distribution with varying arrival rates. All requests utilize greedy sampling for generation.


\begin{table}[h]
    \centering
    \resizebox{0.99\linewidth}{!}{
    \begin{tabular}{c|ccccc}
        \toprule
        Dataset          & Input & (160M, 7B) & (160M, 13B) & (160M, 33B) & (1B, 70B) \\
        \midrule
        ShareGPT         & 227 ($\pm$264)   & 256 ($\pm$249) & 274 ($\pm$255) &  269 ($\pm$233) & 227 ($\pm$208) \\
        Sonnet           & 517  ($\pm$10)   & 141 ($\pm$22) & 145 ($\pm$19) &  139 ($\pm$15) & 135 ($\pm$18) \\
        CNN/Daily Mail   & 1067 ($\pm$537)  & 108 ($\pm$41) & 94 ($\pm$47) &  82 ($\pm$44) & 85 ($\pm$48) \\
        \bottomrule
    \end{tabular}
    }
    \caption{Dataset statistics, input/output length.}
    \label{tab:dataset-stats-1}
    \vspace{-10pt}
\end{table}

\begin{table}[h]
    \centering
    \resizebox{0.75\linewidth}{!}{
    \begin{tabular}{c|cccc}
        \toprule
        Dataset          & (160M, 7B) & (160M, 13B) & (160M, 33B) & (1B, 70B) \\
        \midrule
        ShareGPT         & 0.62 & 0.56 &  0.59 & 0.78 \\
        Sonnet           & 0.53 & 0.62 &  0.65 & 0.92 \\
        CNN/Daily Mail   & 0.54 & 0.56 &  0.61 & 0.83 \\
        \bottomrule
    \end{tabular}
    }
    \caption{Dataset statistics, token acceptance rate.}
    \label{tab:dataset-stats-2}
\end{table}

\begin{table}[t]
    \centering
    \resizebox{0.99\linewidth}{!}{
    \begin{tabular}{rrrrrrrrr}
    \toprule
    (Target, Data)  & QPS & Org  & k=1  & k=3  & k=5  & k=7  & TS \\
    \midrule
    \multirow{4}{*}{\makecell{Vicuna-7B \\ Draft-based \\ ShareGPT}}  
                            &  1  & 2.63   & 1.85   & 1.36        & 1.34  & {\bf 1.31}  & 1.34 \\
                            &  4  & 3.39   & 2.51   & {\bf 1.91}  & 1.92  & 2.05        & 1.90 \\
                            &  8  & 4.86   & 3.78   & {\bf 2.91}  & 3.25  & 4.19        & 2.98 \\
                            & 16  & {\bf 13.73}  & 15.91  & 15.01 & 23.50       & 33.40       & 13.62 \\
    \hline
    \multirow{4}{*}{\makecell{Vicuna-7B \\ Draft-based \\ Sonnet}}                        
                            &  1  & 2.77  & 2.08  & {\bf 1.71}   &  1.76       & 1.84       & 1.74 \\
                            &  4  & 4.36  & 3.36  & {\bf 2.73}   &  3.08       & 3.69       & 2.82 \\
                            &  8  & 13.30  & {\bf 12.81}  & 13.02  &  21.64    & 33.26      & 12.80 \\
                            & 16  & {\bf 72.38}  & 74.74  & 74.4 & 96.07      & 120.74     & 74.03 \\
    \hline
    \multirow{4}{*}{\makecell{Vicuna-7B \\ PLD \\ CNN-DailyMail}}                        
                            &  1  & 2.23   & 2.06   & 1.31        & 1.20   & {\bf 1.18}    & 1.23 \\
                            &  4  & 5.67   & 7.62   & 3.53        & {\bf 3.33}   & 3.50    & 3.25 \\
                            &  8  & 42.28  & 42.41  & 33.77       & {\bf 32.78}  & 34.03   & 31.07 \\
                            & 16  & 138.63 & 141.27 & 120.00      & {\bf 119.97} & 120.70  & 117.01 \\
    \hline
    \multirow{4}{*}{\makecell{Vicuna-13B \\ Draft-based \\ ShareGPT}}                        
                            &  0.5  & 8.43   & 5.51   & 4.09  & {\bf 3.83}        & {\bf 3.83}  & 3.87 \\
                            &  1    & 9.98   & 6.66   & 5.07  & {\bf 4.85}        & 5.05        & 4.94 \\
                            &  2    & 12.90  & 8.66   & 6.43  & {\bf 6.32}        & 6.92        & 6.37 \\
                            &  4    & 34.89  & 20.67  & {\bf 12.70} & 18.34       & 28.00       & 13.31 \\
    \hline
    \multirow{4}{*}{\makecell{Vicuna-13B \\ Draft-based \\ Sonnet}}                        
                            &  0.5  & 6.40  & 3.94  & 2.57  & 2.07      & {\bf 1.96}     & 2.05 \\
                            &  1    & 7.36  & 4.54  & 2.99  & 2.45      & {\bf 2.36}     & 2.5 \\
                            &  2    & 10.35 & 6.08  & 3.88  &  3.19  & {\bf 3.16}        & 3.27 \\
                            &  4    & 45.67 & 24.89 & 9.71  & {\bf 7.20} &  8.97         & 7.60 \\
    \hline
    \multirow{4}{*}{\makecell{Vicuna-13B \\ PLD \\ CNN-DailyMail}}                        
                            &  0.5  & 4.09  & 3.44  &  3.04  &  2.96       &  2.96  & 2.96 \\
                            &  1    & 4.95  & 4.24  &  3.74  & {\bf 3.70}  & {\bf 3.74}  & 3.74 \\
                            &  2    & 8.34  & 7.18  &  6.03  & {\bf 6.01}  & 6.36  & 6.06 \\
                            &  4    & 58.65 & 53.63 & 46.65 & {\bf 46.13} & 49.09  & 46.87 \\
    \bottomrule
    \end{tabular}
    }
    \caption{L40S average request latency in seconds across different datasets and QPS. TS: \sys. PLD: prompt lookup decoding. We highlight the configuration with the lowest latency, comparing both the original and fixed proposed lengths.}
    \label{tab:l40s-performance}
    \vspace{-20pt}
\end{table}

\begin{table}[t]
    \centering
    \resizebox{0.99\linewidth}{!}{
    \begin{tabular}{lrrrrrrrr}
    \toprule
    (Target, Data)  & QPS & Org  & k=1  & k=3  & k=5  & k=7  & TS \\
    \midrule
    \multirow{4}{*}{\makecell{Vicuna-33B \\ Draft-based \\ ShareGPT}}                        
                            &  1  & 2.88  & 2.14  & {\bf 1.73}  & {\bf 1.73}  & 1.81  & 1.73 \\
                            &  4  & 3.73  & 3.06  & {\bf 2.51}  & 2.65  & 3.10  & 2.65 \\
                            &  8  & 4.69  & 4.61  & {\bf 4.24}  & 6.67  & 12.63  & 4.44 \\
                            & 16  & {\bf 11.21}  &  24.31 & 32.82 & 47.29 & 62.22 & 11.67 \\
    \hline
    \multirow{4}{*}{\makecell{Vicuna-33B \\ Draft-based \\ Sonnet}}                        
                            &  1  & 2.37  & 1.66   & 1.20  & 1.10  & {\bf 1.05}  & 1.04 \\
                            &  4  & 3.32  & 2.47   & 1.74  & 1.64  & {\bf 1.60}  & 1.61 \\
                            &  8  & 5.64  & 5.35   & {\bf 4.18}  & 5.35  & 7.77  & 4.21 \\
                            & 16  & {\bf 43.32}  & 49.38 & 49.43 & 57.46 & 68.02 & 46.16\\
    \hline
    \multirow{4}{*}{\makecell{Vicuna-33B \\ PLD \\ CNN-DailyMail}}                        
                            &  1  & 0.62  & 0.48  & 0.39  & {\bf 0.38}  & {\bf 0.38}  & 0.39 \\
                            &  4  & 0.97  & 0.78  & {\bf 0.66}  & {\bf 0.66}  & 0.69  & 0.65 \\
                            &  8  & 1.70  & 1.54  & {\bf 1.36}  & 1.44  & 1.66  & 1.35 \\
                            & 16  & {\bf 28.80}  & 32.21 & 34.17 & 39.45 & 45.17 &  
                            30.87\\
    \hline
    \multirow{4}{*}{\makecell{Llama3-70B \\ Draft-based \\ ShareGPT}}                        
                            &  1  & 3.56  & 2.44  & 1.76  & {\bf 1.69}  & {\bf 1.69}  & 1.69 \\
                            &  4  & 4.77  & 3.52  & 2.48  & {\bf 2.39}  & 2.42        & 2.38 \\
                            &  8  & 5.58  & 4.96  & 3.61  & {\bf 3.58}  & 4.14       & 3.64 \\
                            &  16 & {\bf 15.67} & 23.20 & 21.43 & 25.93 & 31.67       & 16.41 \\
    \hline
    \multirow{4}{*}{\makecell{Llama3-70B \\ Draft-based \\ Sonnet}}                        
                            &  1  & 2.83  & 1.80  & 1.14  & 0.94  & {\bf 0.86}  & 0.86 \\
                            &  4  & 3.63  & 2.47  & 1.55  & 1.29  & {\bf 1.19}  & 1.19 \\
                            &  8  & 5.42  & 4.35  & 2.74  & 2.16  & {\bf 2.09}  & 2.11 \\
                            &  16  & {\bf 18.24}  & 32.48 & 27.84 & 27.37 & 29.18 & 18.60\\
    \hline
    \multirow{4}{*}{\makecell{Llama3-70B \\ PLD \\ CNN-DailyMail}}                        
                            &  1  & 1.51  & 1.16  & 0.91  & 0.84  & {\bf 0.82}  & 0.84 \\
                            &  4  & 2.08  & 1.77  & 1.40  & 1.32  & {\bf 1.31}  & 1.34 \\
                            &  8  & 4.11  & 4.27  & 3.65  & {\bf 3.52}  & 3.78  & 3.58 \\
                            & 16  & {\bf 31.49}  & 44.72 & 45.13 & 48.25 & 51.56 & 32.23 \\
    \bottomrule
    \end{tabular}
    }
    \caption{H100 average request latency in seconds across different datasets and QPS. TS: \sys. PLD: Prompt lookup decoding. We highlight the configuration with the lowest latency, comparing both the original and fixed proposed lengths.}
    \label{tab:h100-performance}
    \vspace{-10pt}
\end{table}

\autoref{tab:l40s-performance} and \autoref{tab:h100-performance} showcase the average request latency of \sys across diverse models (Vicuna-7B/13B/33B and Llama3-70B) and datasets (ShareGPT, Sonnet, CNN-DailyMail) on both L40S and H100 GPUs. The results demonstrate two key aspects of the system's adaptive capabilities:
First, the system intelligently manages speculative decoding under high queries-per-second (QPS) regions. When the QPS is high, the system can automatically disable speculative decoding to maintain optimal performance. For example, on L40S at QPS=16, even minimal speculation (k=1) degrades performance by 14\% (from 13.7s to 15.9s). In such cases, the system reverts to standard decoding, maintaining baseline performance levels (13.6s). 
Second, the results underscore the limitations of fixed proposal lengths and the advantages of dynamic adaptation. While certain configurations appear optimal in specific scenarios (e.g., k=3 for Vicuna-7B achieving 1.4-1.9s latency across various QPS values), these optima don't generalize across model scales. For instance, Vicuna-13B achieves optimal performance with k=3 at QPS=4 (12.7s), but benefits from longer proposals (k=5/7) at lower QPS values (0.5, 1, 2). Similarly, Llama3-70B on the Sonnet dataset, which exhibits high token acceptance rates, shows superior performance with k=7 at low QPS (1, 4, 8).

While \sys generally improves performance, we observe occasional degradation. For example, with Vicuna-33B at QPS=16, the average request latency increases from 43.32s with original decoding to 46.16s with \sys. This overhead stems from additional prefill phases required by \sys compared to standard decoding. As discussed in \autoref{sec:serving-llm-with-dsd}, \sys automatically disables prefilling when batch sizes exceed a certain threshold. The system periodically resets this threshold to prevent permanent deactivation of speculative decoding. Although this reset mechanism can introduce some draft model prefilling overhead when speculative decoding remains disabled during the decoding phase, our experiments show that this overhead is typically minimal. In general, through dynamic adaptation, \sys captures these optimization opportunities that static configurations would miss, consistently achieving latencies that match or approach the optimal proposal length across varying workload conditions.

\begin{figure}
    \centering
    \includegraphics[width=0.7\linewidth]{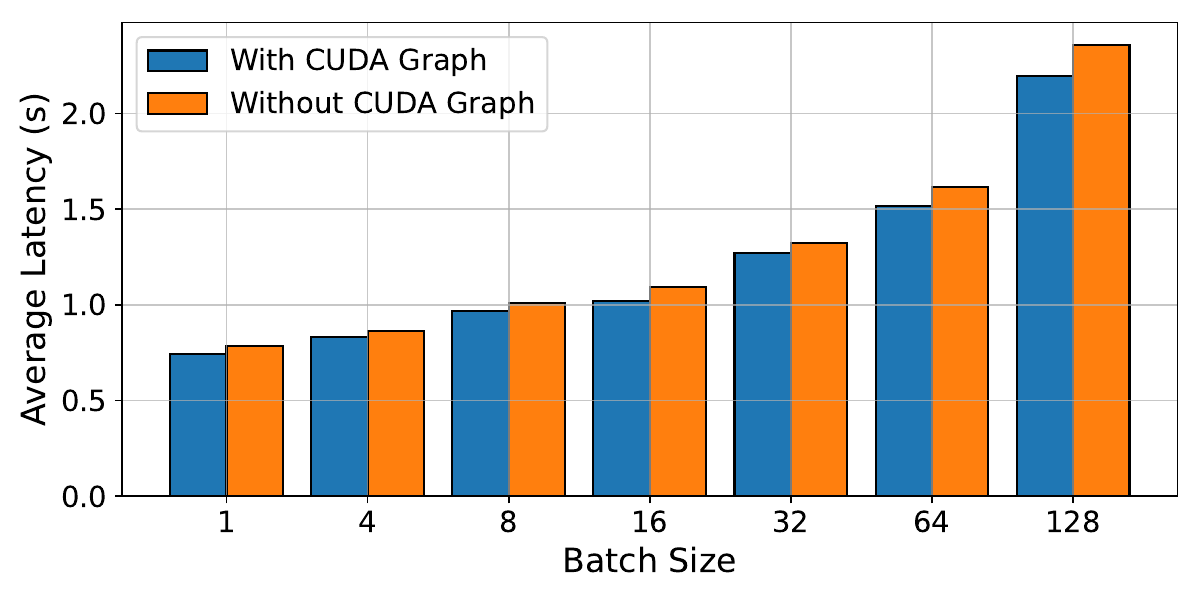}
    \caption{Comparison of average request latency with and without CUDA Graph optimization across various batch sizes.}
    \label{fig:cudagraph}
\end{figure}
\section{Ablation: Effect of CUDA Graph}
Figure~\ref{fig:cudagraph} shows the impact of CUDA Graph optimization on average request latency across different batch sizes. For each batch size, we compare two configurations: one with CUDA Graph enabled (blue bars) and one without CUDA Graph (orange bars). Overall, enabling CUDA Graph consistently reduces latency across all batch sizes, with more pronounced improvements as the batch size increases. The gap between the two configurations widens at larger batch sizes (e.g., 64 and 128), highlighting the growing benefits of CUDA Graph in high-throughput settings. This demonstrates that CUDA Graph is an effective optimization for reducing runtime overhead and improving serving efficiency, particularly under heavy batching scenarios.


    

\end{document}